\documentclass{article}

\usepackage{lineno,hyperref}

\usepackage{arxiv}

\usepackage[utf8]{inputenc} % allow utf-8 input
\usepackage[T1]{fontenc}    % use 8-bit T1 fonts
\usepackage{hyperref}       % hyperlinks
\usepackage{url}            % simple URL typesetting
\usepackage{booktabs}       % professional-quality tables
\usepackage{amsfonts}       % blackboard math symbols
\usepackage{nicefrac}       % compact symbols for 1/2, etc.
\usepackage{microtype}      % microtypography
\usepackage{lipsum}		% Can be removed after putting your text content
\usepackage{graphicx}
\usepackage{doi}
\usepackage[dvipsnames]{xcolor}
\usepackage{subfig}
\usepackage{amsmath}
\usepackage{float}
\usepackage{todonotes}

\title{CASTELO: Clustered Atom Subtypes aidEd Lead Optimization – a combined machine learning and molecular modeling method} 

\author
{Leili Zhang$^{1\ast\dag}$, Giacomo Domeniconi$^{2\ast\dag}$, Chih-Chieh Yang$^1$, Seung-gu Kang$^1$, Ruhong Zhou$^1$, Guojing Cong$^1$\\
\\
\normalsize{$^1$IBM Thomas J Watson Research Center,}\\
\normalsize{1101 Kitchawan Rd, Yorktown Heights, NY 10598}\\
\normalsize{$^2$IBM Research Z\"{u}rich,}\\
\normalsize{S\"{a}umerstrasse 4, 8803 R\"{u}schlikon, Switzerland}\\
\\
\normalsize{$^\dag$Contributed equally to this work;}\\
\normalsize{$^\ast$To whom correspondence should be addressed;}\\
\normalsize{E-mails: zhangle@us.ibm.com, Giacomo.Domeniconi1@ibm.com.}\\
}

\renewcommand{\headeright}{Zhang \textit{et al.} }
\renewcommand{\shorttitle}{CASTELO}

\begin{document}
\maketitle

\begin{abstract}
Drug discovery is a multi-stage process that comprises two costly major steps: pre-clinical research and clinical trials. Among its stages, lead optimization easily consumes more than half of the pre-clinical budget. We propose a combined machine learning and molecular modeling approach that automates lead optimization workflow \textit{in silico}. The initial data collection is achieved with physics-based molecular dynamics (MD) simulation. Contact matrices are calculated as the preliminary features extracted from the simulations. To take advantage of the temporal information from the simulations, we enhanced contact matrices data with temporal dynamism representation, which are then modeled with unsupervised convolutional variational autoencoder (CVAE). Finally, conventional clustering method and CVAE-based clustering method are compared with metrics to rank the submolecular structures and propose potential candidates for lead optimization. With no need for extensive structure-activity relationship database, our method provides new hints for drug modification hotspots which can be used to improve drug efficacy. Our workflow can potentially reduce the lead optimization turnaround time from months/years to days compared with the conventional labor-intensive process and thus can potentially become a valuable tool for medical researchers.
\end{abstract}
%The end-to-end execution starts from MD simulations and produces ranked submolecular structures in the lead molecule. 

% keywords can be removed

\section*{Introduction}
%\begin{itemize}
%    \item lead optimization task, methods and open challenges [Leili]
%    \item MD simulations, motivations and challenges [Leili] 
%    \item ML for MD analysis to overcome the high dimensionality, common ML methods and VAE ones [Giacomo/Leili]
%    \item how to individuate bad/good portion of the molecule to mutate for lead optimization? -> Our subtyping [Leili, maybe before the ML part?]
%    \item Our proposal: Subtyping + CVAE + clustering + comparison to a reference = mutation suggestion [Giacomo]
%    \item ?Description of the Sweetener dataset used as example? [Leili]
%\end{itemize}

At a time of global health crisis, drug discovery is of utter importance to bring the society back to its order. Beyond the crisis, drugs have helped to improve life quality and increase the life expectancy. However, despite the growing research and development expenditure every year \cite{kaitin2010deconstructing, tulum2018innovation}, the yearly FDA-approval of drugs has mostly stalled since 1993 \cite{mullard20202019}. In fact, there were a total of 3,437 FDA approved small-molecule and large-molecule drugs or therapeutics in 2018 \cite{farhadi2018computer}, with a yearly addition of only $\sim$1.2\% (2014-2018 average). The so-called “Eroom’s law” overshadows the pharmaceutical industry, with new molecular entity (NME) drugs per billion dollars spent decreasing ever since 1950 \cite{jones2018biomedical}.  While millions of people globally are suffering from incurable diseases such as Parkinson’s disease \cite{cabreira2019contemporary}, Huntington’s disease \cite{kang2017emerging}, HIV/AIDS \cite{gulick2019long} and hepatitis B \cite{lopatin2019drugs}, technological revolutions are required to find the cures and possibly avert Eroom’s law.

Computer-aided drug discovery (CADD) methods have been widely used in the pharmaceutical industry before preclinical trials, from quantitative structure-activity relationships (QSAR, \cite{gramatica2007principles}), pharmacophore modeling \cite{yang2010pharmacophore} to drug-target docking algorithms \cite{kitchen2004docking}. Along with the conventional high-throughput screening (HTS \cite{macarron2011impact}) method and rising fragment-based drug discovery (FBDD \cite{murray2009rise}) method, computational methods have helped to lower the initial cost of drug discovery (target-to-hit and hit-to-lead, specifically). The resulting lead products usually have a dissociation constant ($K_\text{d}$) in the micromolar range. The subsequent process would aim to optimize the drug lead by lowering the $K_\text{d}$ value to the nanomolar range while ensuring the safety of the drug (and therefore termed “lead optimization”). However, the optimization processes highly rely on domain knowledge and luck because of the astronomical possibilities in the chemical space. Several computational methods have been developed to assist lead optimization. For example, a semi-automated optimization method was suggested by Lewis, combining 3D-QSAR model, random structure permutation methods and human decisions \cite{lewis2005general}. Jain applied a combined QSAR-docking method to an existing library of CDK2 inhibitors and showed encouraging agreement with the experiments \cite{jain2010qmod}. Tang \textit{et al.} improved zinc endopeptidase inhibitor by intuitively suggesting drug modifications from a binding mode suggested by the molecular dynamics (MD) simulation \cite{tang2007computer}. Jorgensen \textit{et al.} adopted Monte Carlo (MC) simulations and free energy perturbation (FEP) simulations to virtually screen chemical groups on a drug lead \cite{jorgensen2006computer}. However, despite these developments, lead optimization remains the most costly step before the clinical trials, constituting $\sim$74\% of the pre-clinical costs \cite{paul2010improve}. A common issue of the previous computational methods lies in the generation of new structures, which mostly relies on inefficient random guesses. Therefore, a systematic way of identifying improvable molecular fragments is needed to assist lead optimization decisions. Notably, computer-aided FBDD algorithms \cite{bian2018computational} can suggest leads by combining the fragment hits (fragments with relatively strong affinity, typically with $K_\text{d}$ $<$ 10 mM \cite{price2017fragment}). But the use is limited once the lead has been identified.

\begin{figure}[t]
\centering
\includegraphics[width = 0.9\textwidth]{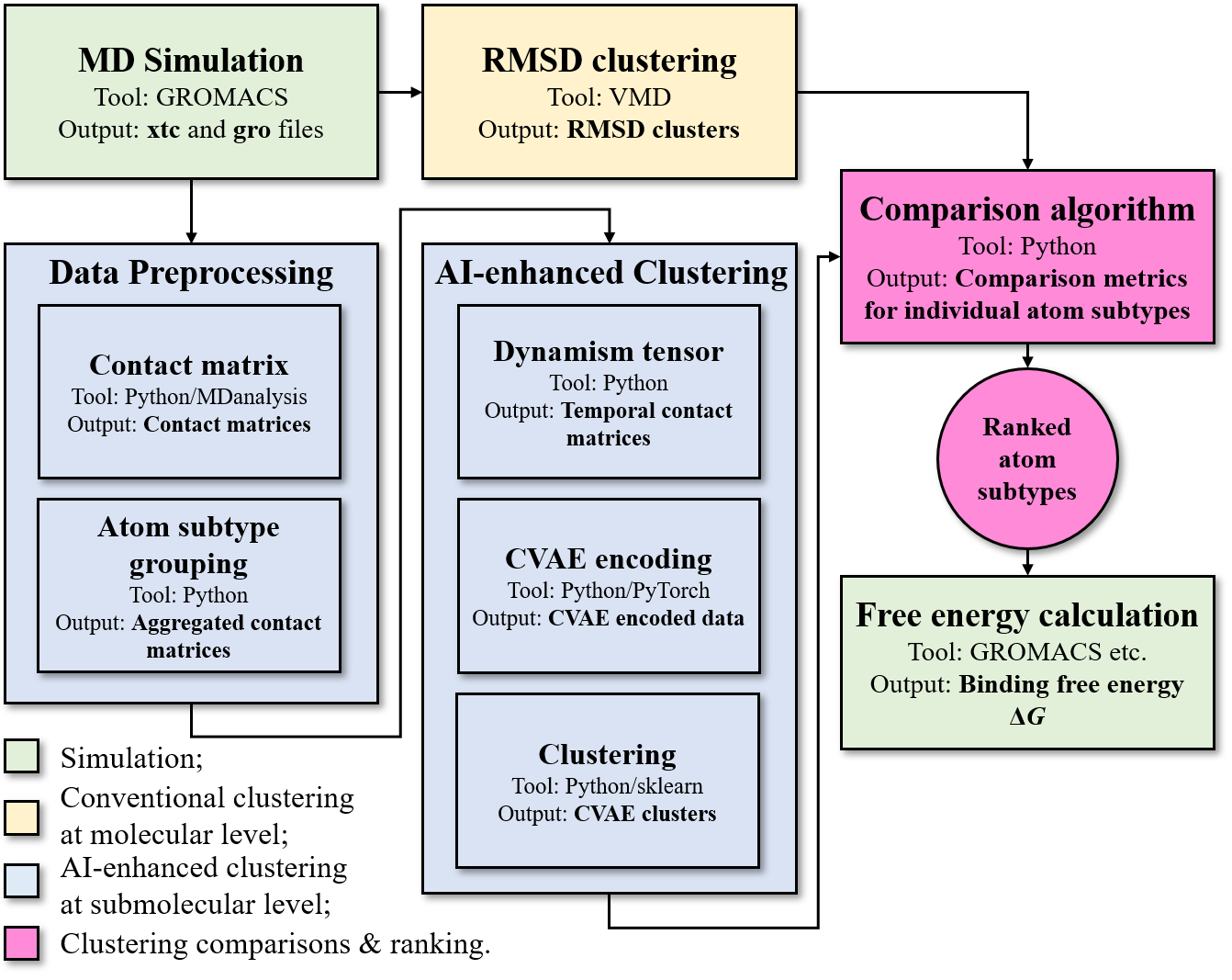}
\caption{The general pipeline for CASTELO. The starting point is the generation of MD trajectories, with tools such as GROMACS. RMSD clustering can be done with VMD software. In another route, we process MD trajectories with python scripts to obtain contact matrices. Atom subtype information is used to aggregate the calculated contact matrices. Following that, dynamism tensors with temporal information is generated on top of the contact matrices using python scripts. CVAE model is used to encode the dynamism data, before clusters are calculated with tools such as HDBSCAN. Finally, we converge the two routes by comparing clusters from conventional RMSD clustering and CVAE clustering with proposed comparison metrics. The atom subtypes are ranked, as the final output of CASTELO. With domain knowledge, we suggest modifications for the lowest ranked atoms. Methods such as free energy perturbation calculations can be used to verify CASTELO's suggestions.}
\label{fig:scheme}
\end{figure}

In this study, we propose a computational method, coined Clustered Atom Subtypes aidEd Lead Optimization (CASTELO), that identifies modifiable submolecular moieties in a lead molecule to narrow down the random guesses to a few possibilities. The process is depicted in Figure \ref{fig:scheme}. Briefly, we first obtain the target-lead binding complexes from either crystal structures or reasonable computational methods (such as homology modeling and MD simulations). Using this structure, $\sim$100 ns of MD simulations are then conducted. Subsequently, contact matrices, containing the relative distances between atoms in the lead and atoms in the target protein, are extracted from the simulations. The contact matrices are processed to contain temporal information (coined as "dynamism tensor") and atom subtype information (see details below).
Compressed vectors are then created with convolutional variational autoencoder (CVAE) using the processed contact matrices. 
CVAE automatically reduces the high dimensionality of the contact matrix into a latent space where states that share similar structural and energetic characteristics \cite{bhowmik2018deep} to each other.
In order to consider the dynamic behavior of the target and the lead, each time step is modeled with a dynamism tensor (a 2-dimensional contact matrix) that contains target-lead interacting information of the MD simulation. This tensor, closely resembling as a 2-channel image, is then fed to the CVAE. 
The latent space representation generated by the CVAE allows the clustering of the time steps of the MD simulation, grouping time steps with similar behavior. Stable snapshots will be clustered in big clusters, while unstable ones will generate a number of small clusters.  

The goal of our method is to discover specific submolecular moieties of the lead molecule that harm the target-lead interactions (thus termed as "malicious atoms"). One way to group the atoms in a molecule to "submolecular moieties" is to categorize them in subtypes based on their physical properties (see Method for details). To differentiate the contributions of the subtypes, CVAE and clustering are not only applied on the whole dynamism tensors, but also to the tensors of each subtype. Clustering information are thus generated for each subtype and compared with conventional clustering information (such as from root mean sqaure displacement (RMSD) clustering) of the whole molecule. Finally, using comparison metrics such as cosine similarity or average difference (see details below), we rank the subtypes from malicious atoms to beneficial atoms. If the overall simulation renders a stable binding structure for the lead (which is often the case), the atom subtypes with lower values of comparison metrics are labeled as malicious atoms, i.e. modifiable atoms for lead optimization, because of their “deviation” from the stable binding state of the rest of the molecule. To verify if these atoms are indeed improvable, we modify the suggested atoms and calculate binding free energy change using FEP calculations. 

%We repeated this procedure for each residue of the antigen. With this approach we were able to compare the overall binding behavior with the behavior of each residue, and then propose antigen residues for favorable mutation. We used a modified dV metric57, mdV, in order to measure the disagreement between the reference clustering (with the whole contact matrix) and each individual residue clustering. Given two clusterings C1 and C2 of a set V with objects u and v in V, dV is computed as follows: 
%We then cluster over the latent space vectors generated by CVAE allowing the discrimination of big clusters with several time steps of similar behavior, i.e. stable situation, or small clusters, i.e. unstable. 
%We used HDBSCAN56 as our clustering methodology. This algorithm overcomes the limitation of knowing in advance the number of clusters, that is the typical drawback of partitioning techniques such as K-means, and also the density threshold, typically required by the standard DBSCAN.
%: the contacts in $t$ and $t-delta$, where $delta$ is a parameter of the framework. 
%Finally, we compare the CVAE clustering information with VMD clustering information using comparison metrics such as cosine similarity. The atom subtypes with lower cosine similarity are labeled as modifiable atoms, because of their “deviation” from the binding state of the rest of the molecule. To verify if these atoms are indeed improvable, we modify the suggested atoms and calculate binding free energy change using FEP simulations.

\begin{figure}[h]
\includegraphics[width = 0.9\textwidth]{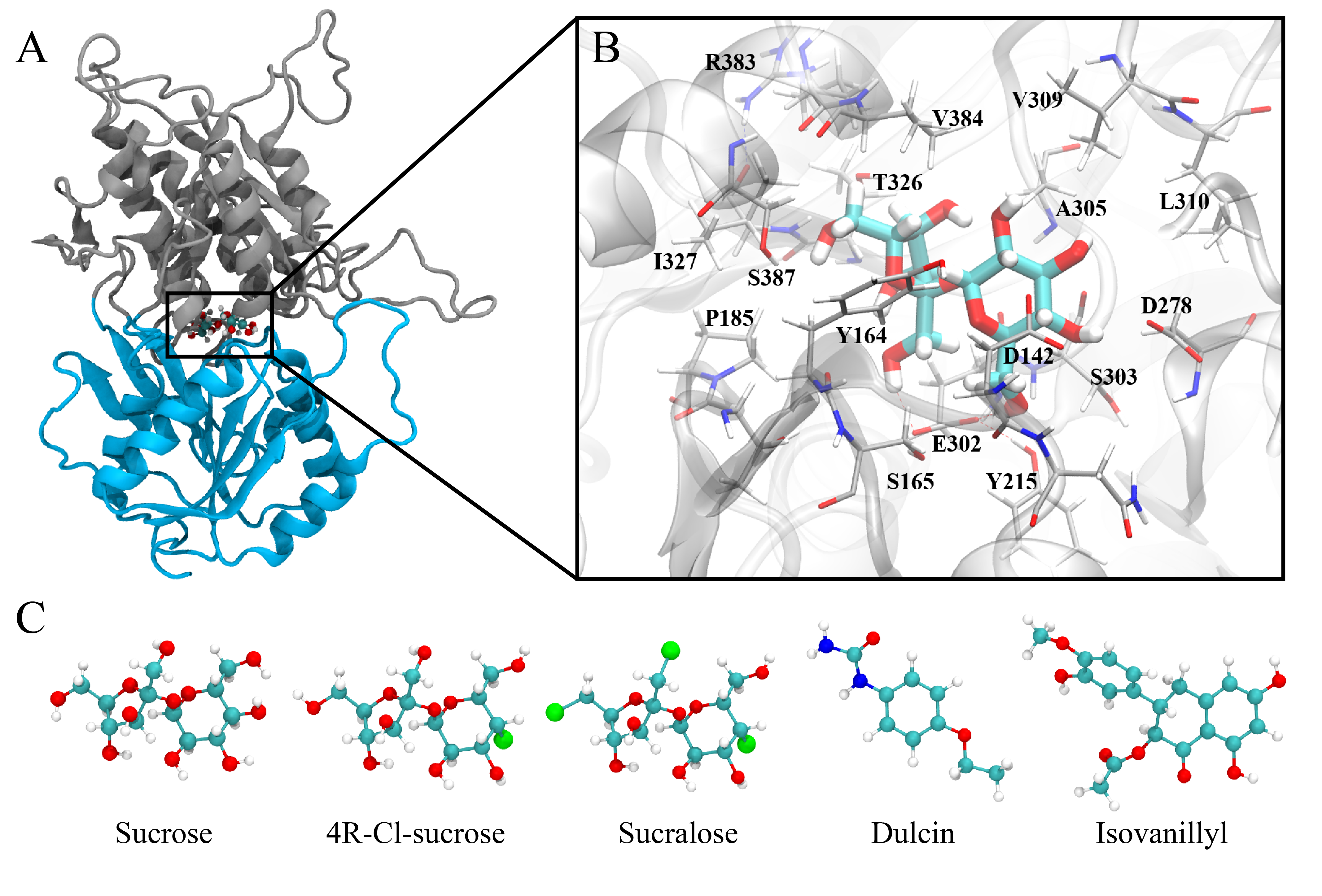}
\caption{The structure of sucrose bound T1R2 is shown in (A). The detailed interactions between sucrose and its surrounding residues are depicted in (B). The structures of the five sweeteners tested in this study are illustrated in (C).}
\label{fig:sweet}
\end{figure}

As an example, we applied this process to the sweetness taste receptor T1R2 with five well-known sweeteners: sucrose (the reference for relative sweetness, RS = 1 \cite{supersweet}), (1R,2R,3R,4R,5R)-4-Chloro-1-[(2R,3S,4S,5R)-3,4-dihydroxy-2,5-bis(hydroxymethyl)oxolan-2-yl]oxy-6-(hydroxymethyl)oxane-3,4-diol (4R-Cl-sucrose thereafter, RS = 5 \cite{hough1978intensification}), sucralose (RS = 600 \cite{supersweet}), dulcin (RS = 250 \cite{supersweet}), and an isovanillyl sweetener (isovanillyl thereafter, RS = 400 \cite{bassoli2002isovanillyl}). The sweetness taste receptor structure is taken from Perez-Aguilar et al. \cite{perez2019modeling}, shown in Figure \ref{fig:sweet}A. The sweetener molecules are shown in Figure \ref{fig:sweet}C. The search of the binding modes for all five sweeteners follows the the combined docking-MD method reported previously \cite{zhuang2018binding}. With CASTELO, we are able to suggest modifications that improve molecular sweetness of the simulated molecules by identifying malicious atoms. Importantly, no structure-activity relationship data are needed for this process. The minimal requirement is a working lead molecule and a target protein structure. %Furthermore, we applied CASTELO to SARS-CoV-2 RdRp and found plausible suggestions on modifiable atoms with the same process and same hyperparameters.

%\section*{Related Work}
%Text and results for this section...

\section*{Methods}
\label{sec:methods}

\subsection*{Molecular docking}
The protein structure of T1R2 was taken from Perez-Aguilar et al. \cite{perez2019modeling}. The structures of sucrose, 4R-Cl-sucrose, sucralose, dulcin and isovanillyl were constructed with Jmol. \cite{hanson2010jmol} We refered to previous studies \cite{zhang2010molecular} and assumed that the "flytrap" domain (referring to the boxed region in Figure \ref{fig:sweet}A due to the resemblance between T1R2 and the venus flytrap) of T1R2 should be the binding domain for sweeteners. Initial binding structures were then constructed by selecting the "flytrap" domain in T1R2 using Autodock Vina. \cite{trott2010autodock}

\subsection*{Molecular dynamics simulation (dataset)}
The initial binding structures constructed from Autodock Vina were solvated with water in a simulation box of 10${\times}$10${\times}$10 nm$^3$. 0.1 M of NaCl was then used to ionize and neutralize the simulation box. The resulting simulation systems contain roughly 105,000 atoms. CHARMM36 force field \cite{huang2013charmm36} was used for the T1R2 protein, water and ions. CGenFF force field was generated with the online server \cite{vanommeslaeghe2012automation} for the five sweeteners. NAMD2.9 \cite{phillips2005scalable} was used to run all simulations on Blue Gene supercomputer. Particle mesh Ewald (PME) was used for electrostatic interaction calculations with a grid size of 1 {\AA}. A switching function was used for van der Waals (VDW) interactions calculations, where switching distance was 10 {\AA}, cutoff distance was 12 {\AA} and VDW pairlist distance was 13.5 {\AA}. Before production runs, 20,000 steps of energy minimization and 250 ps of equilibration using 0.5 fs time step were performed. For the production runs we used 2 fs as the time step. Depending on the stability of the structures, 50 to 100 ns of simulations were performed for each of the docked structures. Snapshots of the trajectories were saved every 20 ps for the data collection.

\subsection*{Free energy perturbation (FEP)}
Following previous studies \cite{zhuang2018binding, zhang2020structural}, we used Zwanzig equation \cite{zwanzig1954high} and thermodynamic cycle shown in Figure \ref{fig:thermocycle} to calculate the relative binding free energy of the five sweeteners with "muted ethane" (refered to as ethane thereafter for simplicity) as the reference point. Muted ethane had the structure of ethane but was chargeless. We performed two sets of simulations with NAMD2.9 for each sweeteners (the thermodynamic cycle is shown in Figure \ref{fig:thermocycle}): the "bind" simulations where T1R2 was present and sweeteners were mutated to ethane; and "free" simulations where T1R2 was absent and sweeteners were mutated to ethane. The resulting free energy change was denoted as ${\Delta}F_{bind}$ and ${\Delta}F_{free}$. The relative binding free energy of the sweeteners can then be calculated as: ${\Delta\Delta}F_{SWT} = {\Delta}F_{SWT} - {\Delta}F_{ethane} = {\Delta}F_{free} - {\Delta}F_{bind}$. If we assume that the relative sweetness of the sweeteners was a direct measurement of the dissociation constant $K_d$, then we can compute the relative sweetness as follows:
\begin{equation} \label{eq:rs}
CRS(SWT) = e ^ {- ({\Delta\Delta}F_{SWT} - {\Delta\Delta}F_{sucrose}) / RT }
\end{equation}

\subsection*{Grouping submolecular moieties using atoms subtypes}
\label{sec:subtyping}
We categorized atoms using the VDW parameters of the atom types (such as CG331) from the CHARMM36 force field. Atom types that fell in the range of 10\% of $\sigma$ (particle size) and $\epsilon$ (dispersion energy) with each other were classified into the same "atom subtype". Such classification resulted in 46 atom subtypes from the approximate 300 atom types from the CHARMM36 protein, carbohydrate and cgenff force field. The 46 atom subtypes were used to differentiate "beneficial atoms" (atom subtypes that likely strengthened sweetener-T1R2 interactions) from "malicious atoms" (atom subtypes that likely weakened sweetener-T1R2 interactions). Note that we chose these the atom subtypes based on our experience with the current dataset. Other subtyping method was not excluded in our patent application and future studies.

\subsection*{AI-enhanced Clustering} %Convolutional variational autoencoder (CVAE) and Clustering
\label{sec:cvae}

\begin{figure}[t]
\includegraphics[scale=0.5]{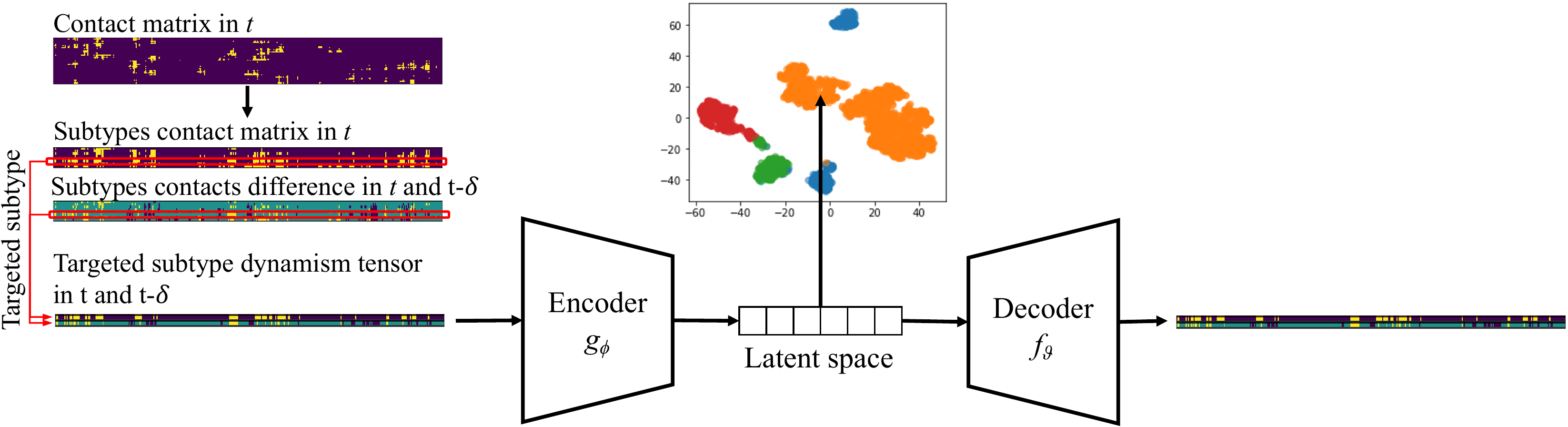}
\caption{CVAE pipeline. Left part of the figure shows the input preparation: the contact matrix of a time step $t$ is grouped over the $y$-axis by drug atom subtypes (yellow cells are contacts). Each contact matrix of a time step $t$ is paired to its temporal difference with a previous one $t-\delta$ to form a dynamism tensor (violet/yellow cells are new/disappearing contacts, green cells are stable). 
Each atom subtype is selected in turn and fed to the CVAE. The representation in the latent space is used to cluster the time steps by similar behaviors.}
\label{fig:cvae}
\end{figure}

Autoencoders are unsupervised neural networks widely used for dimensionality reduction and pattern recognition. Their architecture is composed of an encoder part $g_\phi$ that compresses an input into a latent space and a decoder part $f_\theta$ that reconstructs the original input using the low-dimensional features.
Variational autoencoders (VAEs) introduce the optimization constraint to the latent space to be normally distributed, this coerces the network to distribute the information more evenly into the latent space \cite{doersch2016tutorial}. 
In MD simulation analysis, a common input is a contact matrix from a single MD time step; hence, convolutional layers (CVAE) rather than regular feedforward are typically used. This leads to filter maps that can better recognize local patterns independently of the position. 

We used CVAE to model each MD trajectory in a low-dimensional space. %For the ML analysis, MD trajectories consisted of 2500 frames saved every 200ps, for a total of 500ns simulation. 
Similar to previous works \cite{tsuchiya2019autoencoder,bhowmik2018deep}, we model the input as a contact matrix between atoms of the drug molecule and atoms of the protein in each structure of the simulation.
Specifically, the contact matrix has size of $N \times M$, where N is the number of drug atoms and M is the number of protein atoms. 
Our goal is to find binding patterns during the simulation, hence the temporal information over time is important to locate stable and unstable states. We enriched each time step $t$ input with a contacts' dynamism  representation: the difference of contacts between the current time $t$ and a previous one $t-\delta$, where $\delta$ is a parameter of our framework. To maintain the the dynamism matrix binary as well as independent to the direction of the movement, we use the absolute value of the difference. In this matrix, 0 cells represents a stable situation, while 1s show a dynamic behavior (either a new contact or the disappearance of a previous contact).
Each time step of the simulation is then modeled as a tensor of $2 \times N \times M$ (termed as the dynamism tensor), with the first two dimensions representing the contact matrix in $t$ and the temporal difference of the contacts in $t$ and $t-\delta$. We used $\delta=500$ (i.e. 10 ns) in our experimental setting. 

CVAE represents each contact tensor in a $d$-dimensional latent space, i.e. a vector of $d$ elements. This intrinsically brings time steps with similar contact matrices to closer positions in the latent space, allowing clustering techniques to group sets of time steps with similar behavioral patterns. In other words, applying clustering over the latent space vectors allows the discrimination of big clusters with several time steps of similar behavior, i.e. stable situation, or small clusters, i.e. unstable. 
In CASTELO, we used HDBSCAN \cite{campello2013density} as our clustering method. This algorithm overcomes the limitation of knowing in advance the number of clusters, that is the typical drawback of partitioning techniques such as K-means, and also the density threshold, typically required by the standard DBSCAN.

The procedure described above provides \textit{general} binding information of a MD simulation, i.e. states of the whole drug molecule. CASTELO provides a fine-grained perspective of the drug molecule behavior: a specific binding view of each atom, or group of atoms. 
At this purpose, we group the drug atoms according to their physical properties (as described in previous Section) and we apply the CVAE and clustering pipeline described above only focusing on the contacts of each individual atom subtype. 
In this case, each time step is sampled by a tensor of size $2 \times 1 \times M$, and squeezing the 1-dimension becomes a 2-channel image of size $2 \times M$. 
Figure \ref{fig:cvae} shows the pipeline with the input preparation of one targeted atom subtype. 
We repeat the CVAE and clustering procedure for each atom subtype of the drug. This approach allows the comparison of the overall binding behavior with that of each type (detailed in next section). 

CASTELO pipeline is completely formed by unsupervised techniques, that are hard to tune and evaluate. In order to provide reliable suggestions, we train slightly different CVAE architectures varying the hyper-parameters. In our experiments we varied the latent dimensions $d$ and the number of convolutional filters $f$. This choice was made because we noticed few cases of all non-clustered samples in cases with latent space dimension $d$ too big. We also tested the results by varying the time $\delta$ of the contacts' dynamism, but the contribution was not pronounced.

\subsection*{Experimental settings} 

As descrived above, we trained a number of CVAE models for each atom subtype. Each input sample, i.e. a time step $t$, is represented by a $2 \times M$ tensor selecting the contact and dynamism vectors related to the targeted subtype. 
In our experimental settings the CVAE encoder part is formed by four convolutional layers with $f$ filters of size 1x7, we decided to keep  the convolutions between contacts and dynamism separated, thus the 1-sized filter in the first dimension. We trained models varying the filter numbers $f \in \{32,64\}$.
Considering the highly narrowed shape of the input, we used a stride of 2 in the second dimension, i.e. on the protein atoms. We did not used pooling or padding. The decoder part is specular to the encoder. We used three different values for the latent space size: $d \in \{3,5,10\}$. For each atom subtype we thus train $A=6$ different CVAE architectures, varying $f$ and $d$.
We combined BCE and KLE as loss function \cite{kingma2013auto}, using RMSProp as optimizer and a learning rate of $0.005$. We trained each CVAE model for a maximum of 600 epochs with an early stopping mechanism after 10 epochs without improvement in the loss. 
We set HDBSCAN with a minimum cluster size of 50, all the other parameters are set to the default values. 

\subsection*{Submolecular moiety suggestion for lead optimization}

The suggestion of malicious atoms to increase the binding affinity of the molecule is made by a comparison of the clustering result of each atom subtype with respect to the clusters obtained while considering the whole molecule.
In this comparison, a clustering that considers the whole molecule is needed as reference. A CVAE model that has  the contact matrices with all the drug atoms as input may be used as reference. Otherwise, in order to avoid possible similarity biases due to the same architecture used by the compared clusters, any other traditional clustering methodology can be used as reference. In our experiments, we decided to use the default RMSD clustering provided by VMD1.9.3. \cite{humphrey1996vmd}

As described in the previous section, each atom subtype has a number of clustering results, one for each trained CVAE architecture. The final comparison is made with the averaged values of the comparison of each architecture with the whole molecule. This comparison also provides a standard deviation value that may be used as agreement score of the different CVAE models of the same atom subtype. I.e. averaged comparison values with small standard deviation indicate a well grounded suggestion. 

More specifically, two types of comparison metrics were explored in this study, with the first one being the cosine similarity (CosSim):
\begin{equation} \label{eq:cossim}
CosSim = \frac{ \sum\limits_{t \in T} C_{t,A} \cdot C_{t,S} }{ \sqrt{ \sum\limits_{t \in T} C_{t,A}^2 } \cdot \sqrt{ \sum\limits_{t \in T} C_{t,S}^2 } }
\end{equation}
where $t$ is a time frame that belongs to the trajectory ($T$), $C_{t,A}$ is the cluster size of the cluster at time $t$ for atom subtype A, calculated with the AI-enhanced clusering, $C_{t,S}$ is the cluster size of the cluster at time $t$ for the whole molecule (S), calculated with conventional methods such as RMSD clustering. Note that CosSim has a range of $[0,1]$, with lower values indicating deviation between the two arrays and higher values indicating similarity. 

The second comparison metric examined was average difference (AvgDiff):  
\begin{equation} \label{eq:avgdiff}
AvgDiff = \frac{ \sum\limits_{t \in T} ( C_{t,A} - C_{t,S} ) }{ \text{count}(T) }
\end{equation}
where $\text{count}(T)$ is the total number of time frames for the trajectory. The values of AvgDiff have a wider range than CosSim, depending on the cluster sizes. Generally speaking, a negative AvgDiff indicates that subtype A is less stable than the whole molecule (S), while a positive AvgDiff indicates the opposite.

After assigning a comparison metric value, each of the atom subtypes is given a final ranking, with the lowest ranked subtype as the suggestion for lead optimization. As a follow-up in this study, we used domain knowledge to suggest some modifications and tested the modifications with FEP calculations (see results for more details).

\section*{Results and discussion}

\begin{table}[t]
\centering
\begin{tabular}{l | c | c | c}
\hline
Sweetener & log(RS)* & ${\Delta\Delta}F$** & log(CRS)*** \\
\hline
Sucrose & 0\cite{supersweet} & -6.9 ${\pm}$ 0.7 & 0 ${\pm}$ 0 \\
4R-Cl-sucrose & 0.70\cite{hough1978intensification} & -10.2 ${\pm}$ 0.8 & 2.33 ${\pm}$ 0.76 \\
Sucralose & 2.78\cite{supersweet} & -11.7 ${\pm}$ 0.8 & 3.38 ${\pm}$ 0.76 \\
Dulcin & 2.40\cite{supersweet} & -10.6 ${\pm}$ 0.4 & 2.61 ${\pm}$ 0.59 \\
Isovanillyl & 2.60\cite{bassoli2002isovanillyl} & -11.1 ${\pm}$ 0.8 & 2.96 ${\pm}$ 0.74 \\
\hline
\end{tabular}
\caption{Free energy. *Relative sweetness. **Binding free energy relative to ethane (kcal/mol). ***Computed relative sweetness is calculated with Equation \ref{eq:rs} in reference to sucrose.}
\label{tab:sweetfe}
\end{table}

\subsection*{RMSD clustering identified stable binding mode and binding motifs}

MD simulations were used to identify stable binding modes of the sweeteners in the binding pocket of T1R2 flytrap domain (shortened for T1R2 thereafter) of human sweetness taste receptor \cite{zhang2010molecular}. We prepared the initial guesses of the binding structures using molecular docking software Autodock Vina \cite{trott2010autodock}.  Around five highest scored docking result(s) were taken as the starting structures in molecular dynamics simulations. Upon the completion of 50-100 ns of simulations, binding mode clusters were classified using RMSD clustering method provided in VMD \cite{humphrey1996vmd} using 2.0 {\AA} as the cutoff. RMSD values were calculated on the sweetener molecules only, while T1R2 structures were aligned. The identified clusters (binding modes) were considered as "stable" if the cluster size persisted longer than 50 ns, following the practices in previous studies \cite{zhuang2018binding, zhang2020structural}. The persistent time of the clusters for all five sweeteners were plotted in Figure \ref{fig:clussize} (only the trajectories with the largest clusters were shown), indicating that stable binding modes were found for each of the sweeteners.

In Figure \ref{fig:sweet}A, we illustrated the stable binding mode extracted from the T1R2-sucrose simulation. The sucrose molecule resided well within the binding pocket popularly referred to as the "venus flytrap". A closer look (Figure \ref{fig:sweet}B) revealed that the crucial binding residues are mostly hydrophilic, including the hydrogen bond receptors D142 D278, E302, among other residues such as Y164, S165, P185, Y215, S303, A305, V309, L310, T326, I327, R383, V384 and S387. If the two loop domains on the right of the boxed region in Figure \ref{fig:sweet}A were considered as the "lips" of the flytrap, we noticed that sucrose bound deeply in the "mouth". One indication of the sweetener being closer to the "lips" would be lower residue numbers among the interacting residues (see Figure \ref{fig:structures}A for a reference of residue numbering). These findings agreed well with the previous binding structure reported by Zhang \textit{et al.} \cite{zhang2010molecular}, where the binding pocket facilitated hydrogen bonds between the 8 hydroxyl groups in sucrose and hydrophilic residues in T1R2. Such hydroxyl group-facilitated sweetness was suggested more than half a century ago and has remained a heavily investigated topic \cite{lindley1976sweetness,suami1992molecular,bruni2018hydrogen}. 

The stable binding structures of the rest of the five sweeteners were illustrated in Figure \ref{fig:structures}. We found that the crucial binding residues for 4R-Cl-sucrose included D142, E302, Y103, N143, Y164, S165, A166, T184, Y215, S303, W304, I306, V309, L310, I325, T326, I327, R383, and S387, highly similar to those of sucrose. This indicated that the binding mechanism mainly remained the same after changing one hydroxyl group of the sucrose to the chlorine atom in 4R-Cl-sucrose. In contrast, the crucial binding residues for sucralose (D142, E302, L41, I67, L71, S144, Y164, S165, A166, I167, T184, H190, S303, A305, V309, and V384) included more residues on the slightly more hydrophobic lips region of the flytrap. This shift was expected because 3 relatively more hydrophilic hydroxyl groups of the sucrose were modified to 3 chlorine atoms in sucralose. Nevertheless, D142 and E302 remained crucial as hydrogen bond receptors for sucralose. 

The stable binding structure of dulcin interacted with mostly hydrophilic residues L41, Y103, D142, Y164, S165, T184, P185, H190, E302, S303, A305, I306, T326, I327, R383, V384, S387, and L448, similar to those of sucrose and 4R-Cl-sucrose. The main hydrogen bonds appeared to be between the urea group of dulcin and D142. Finally, the stable binding structure of isovanillyl interacted with F39, S40, L41, V64, I67, L71, Y103, D142, Y164, S165, Y215, E302, S303, V309, V384, V385, S387, V388, indicating a binding domain that was more hydrophobic and similar to sucralose. Only occasional (weak) hydrogen bonds were found between isovanillyl and E302, along with backbone carbonyl groups of several residues.

\subsection*{Binding free energy of the sweeteners agreed well with experimental relative sweetness}
Starting from the stable binding structures of the five sweeteners in this study, we performed FEP calculations for each by constructing "dual topologies" between muted ethane and the corresponding sweetener (see Methods for more details). The computed binding free energy ${\Delta\Delta}F$ was therefore the relative binding free energy between the sweetener and muted ethane in T1R2 binding pocket. We assumed that muted ethane had a flat binding energy surface in the T1R2 binding site due to its chargeless nature. Then the calculated binding free energy should be relatively accurate among the five sweeteners. Note that similar approaches have been verified in a number of previous publications \cite{zhuang2018binding,zhang2020structural,miyamoto1993absolute,shivakumar2010prediction,williams2018free}. 

We noticed that the general trend of ${\Delta\Delta}F$ values and experimental values of $RS$ agreed well with each other (Table \ref{tab:sweetfe}). The binding mechanisms of the five sweeteners were, however, vastly different as suggested by the interacting residues listed above. To bridge between our speculations on the structural information and quantitation, we reported the electrostatic and VDW contributions in ${\Delta\Delta}F$ by performing energy decomposition for all five FEP calculations (see Table \ref{tab:fedecomp}). We found that sucrose, 4R-Cl-sucrose, and dulcin mostly utilized polar interactions. In contrast, isovanillyl mostly utilized non-polar interactions, agreeing well with the structural speculations. Surprisingly, despite being in a similar binding domain to isovanillyl, sucralose still utilized mostly polar interactions. Comparing the binding modes between sucrose, 4R-Cl-sucrose and sucralose, it was likely that the replacement of hydroxyl groups to chlorine lowered the penalty of the remaining hydrophilic hydroxyl groups in hydrophobic environment, rather than directly boosting the hydrophobic interactions.

We further computed the relative sweetness ($CRS$), listed in Table \ref{tab:sweetfe}. The computed results agreed generally well with the experimental $RS$ values. Such agreement showcased that FEP method could be used to calculate the $RS$ of artificial sweeteners. In later sections, we used FEP calculations to validate that "lead optimization" (meaning to increase the $RS$ value) of the sweetener was possible using CASTELO.

\subsection*{AI-enhanced clustering with all atoms agree well with the RMSD clustering}

\begin{figure}[h]
\includegraphics[scale=0.5]{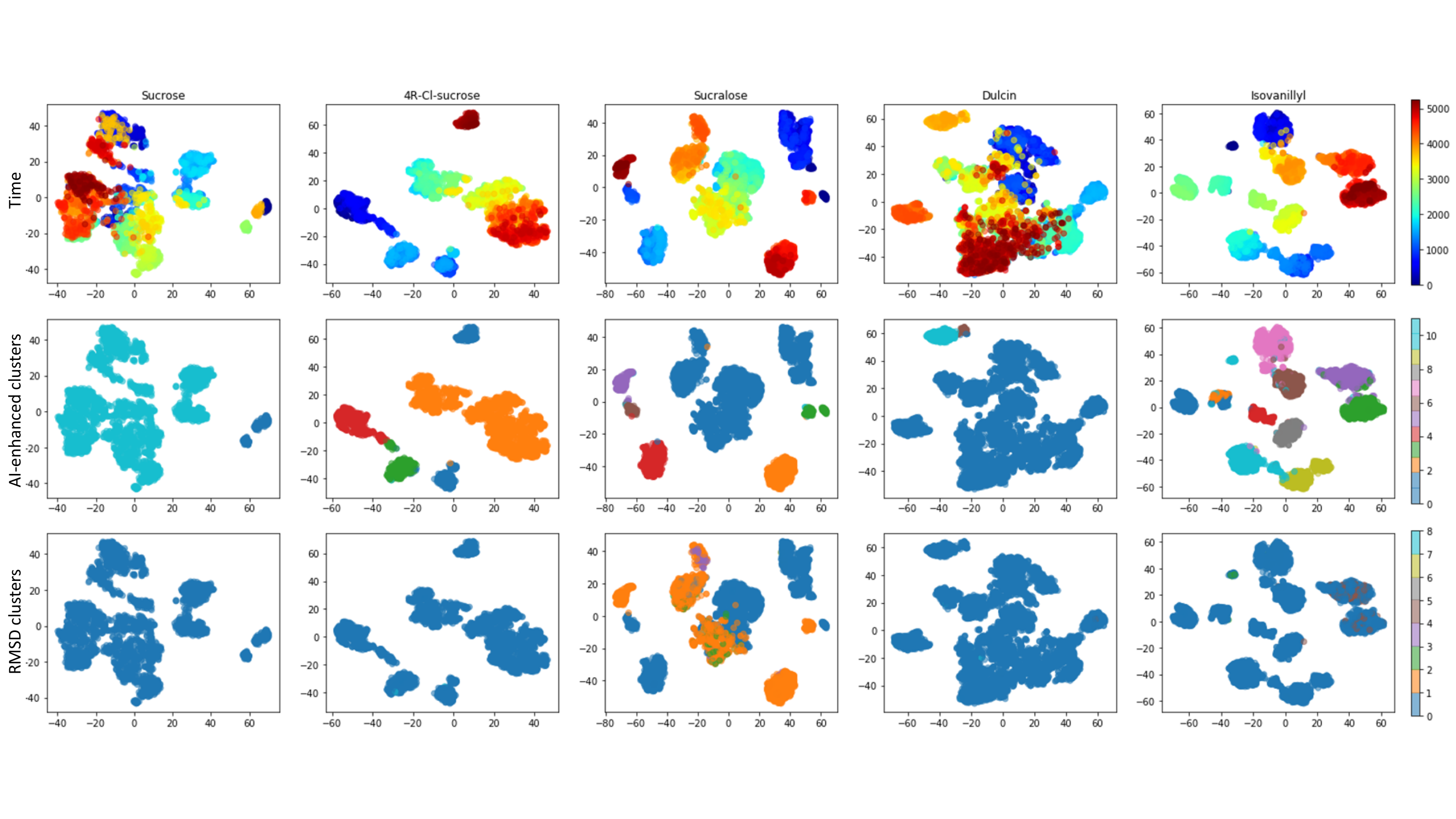}
\caption{T-SNE projection of the CVAE latent representations. The top row uses a color map based on the simulation time. The middle and bottom rows use colors to distinguish clustering results respectively of our AI-enhanced model and of the RMSD. Here we depict the encoded representations generated by the CVAE model with $f=32$ filters and $d=5$ latent dimensions.}

\label{fig:tsne}
\end{figure}

We firstly tested the reliability of our AI-enhanced clustering with a direct comparison to a reference clustering using RMSD in VMD1.9.3. 
In order to get a real comparison, we trained the CVAE models using all the sweetener atoms, i.e. with the same contact matrices used by RMSD as input. To have a 2-dimensional tensor as input of the CVAE, we concatenated the contacts and dynamism tensors in the first dimension, forming a $2N \times M$ input.  
We noticed that the obtained clustering results agreed well with those obtained by RMSD. Figure \ref{fig:tsne} shows the t-SNE \cite{hinton2002stochastic} visualization of the latent representations by the CVAE models of each sweetener, each data point is a time step of the simulation. In the top row we used a color map based on the simulation time. The middle and bottom rows used colors to distinguish clusters found respectively by our AI-enhanced model and by the RMSD. 
%From this figure we can notice several insights.
Firstly, we noted that the HDBSCAN behaved well in clustering time steps projected into the latent space by the CVAE model, this was noticeable by the agreement of colors mapping in the middle row and actual clusters of points. It was also noticeable that, generally, clusters are formed by time frames close in the simulation. This supported the use of the dynamism tensor, that can help the model to distinguish time steps with similar contact matrices but different dynamism. 
The general agreement between the middle and bottom rows proved the reliability of our AI-enhance clustering methodology, allowing us to proceed in a fine-grained analysis of the drug molecule behavior.

\subsection*{AI-enhanced clustering varies with individual atom subtypes, hinting malicious atom subtypes} %(that may weeken protein-sweetener interactions)}

\begin{figure}[h]
\centering
\subfloat{\includegraphics[width=0.33\textwidth]{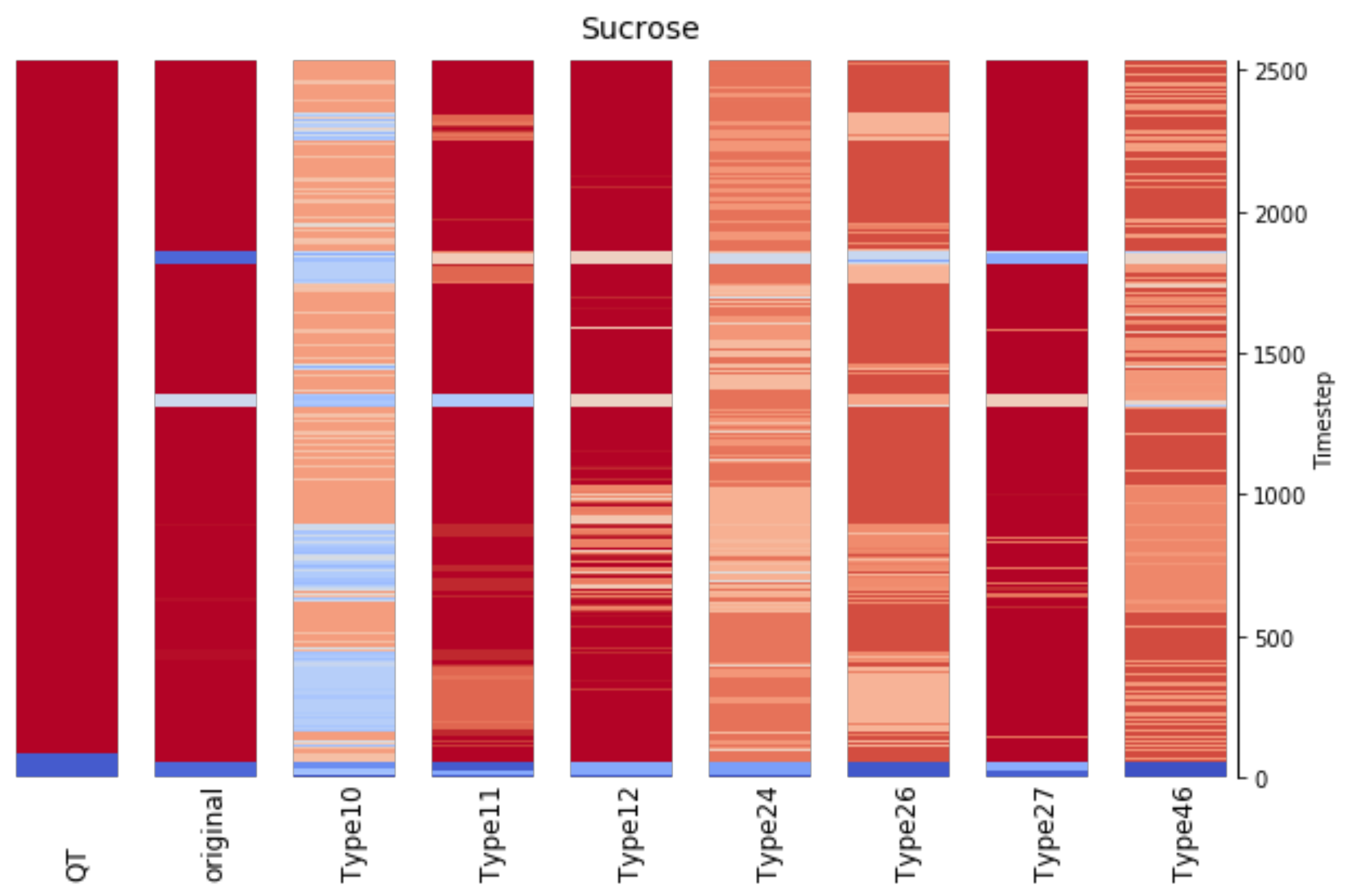}} 
\subfloat{\includegraphics[width=0.33\textwidth]{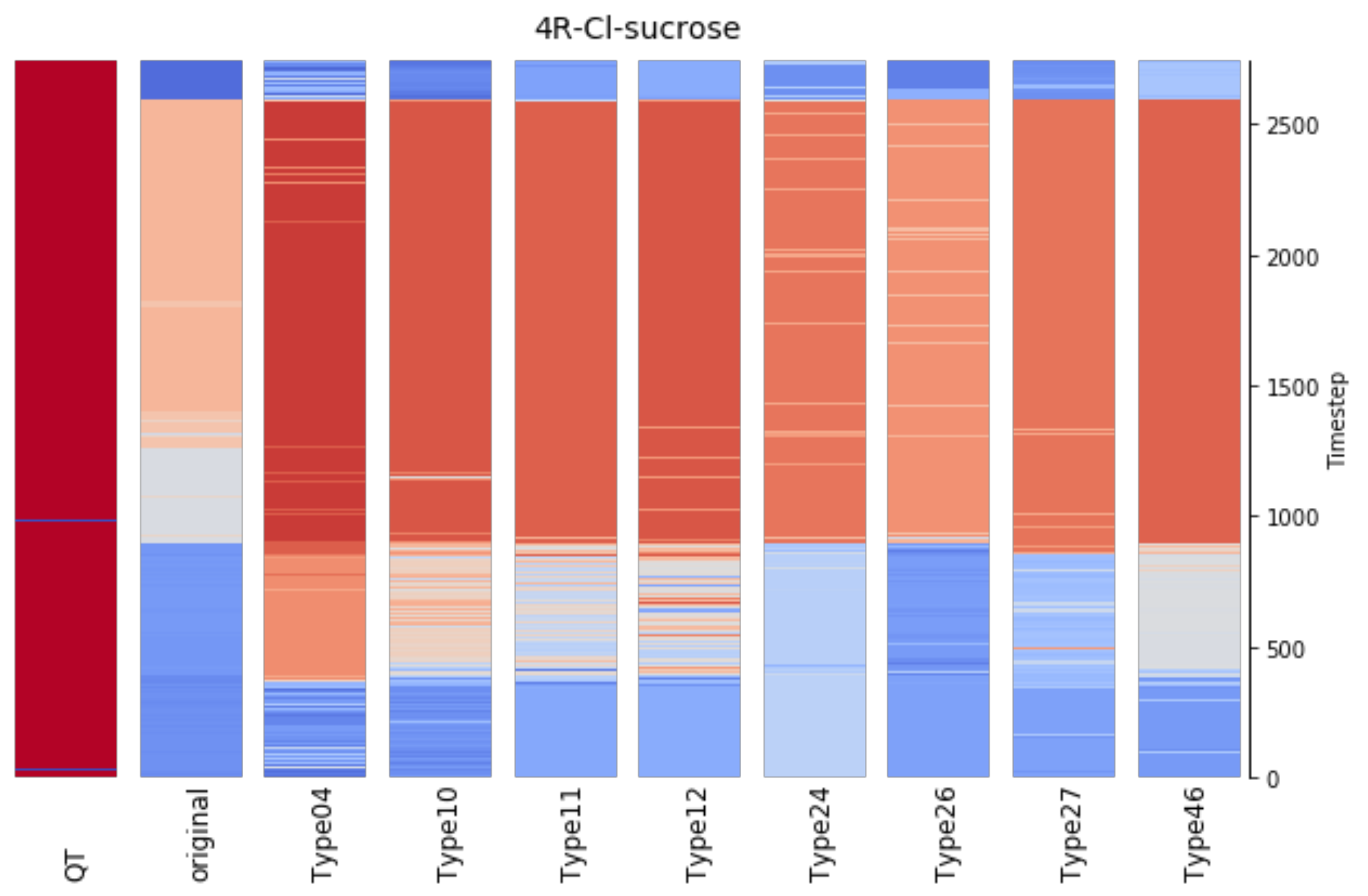}}
\subfloat{\includegraphics[width=0.33\textwidth]{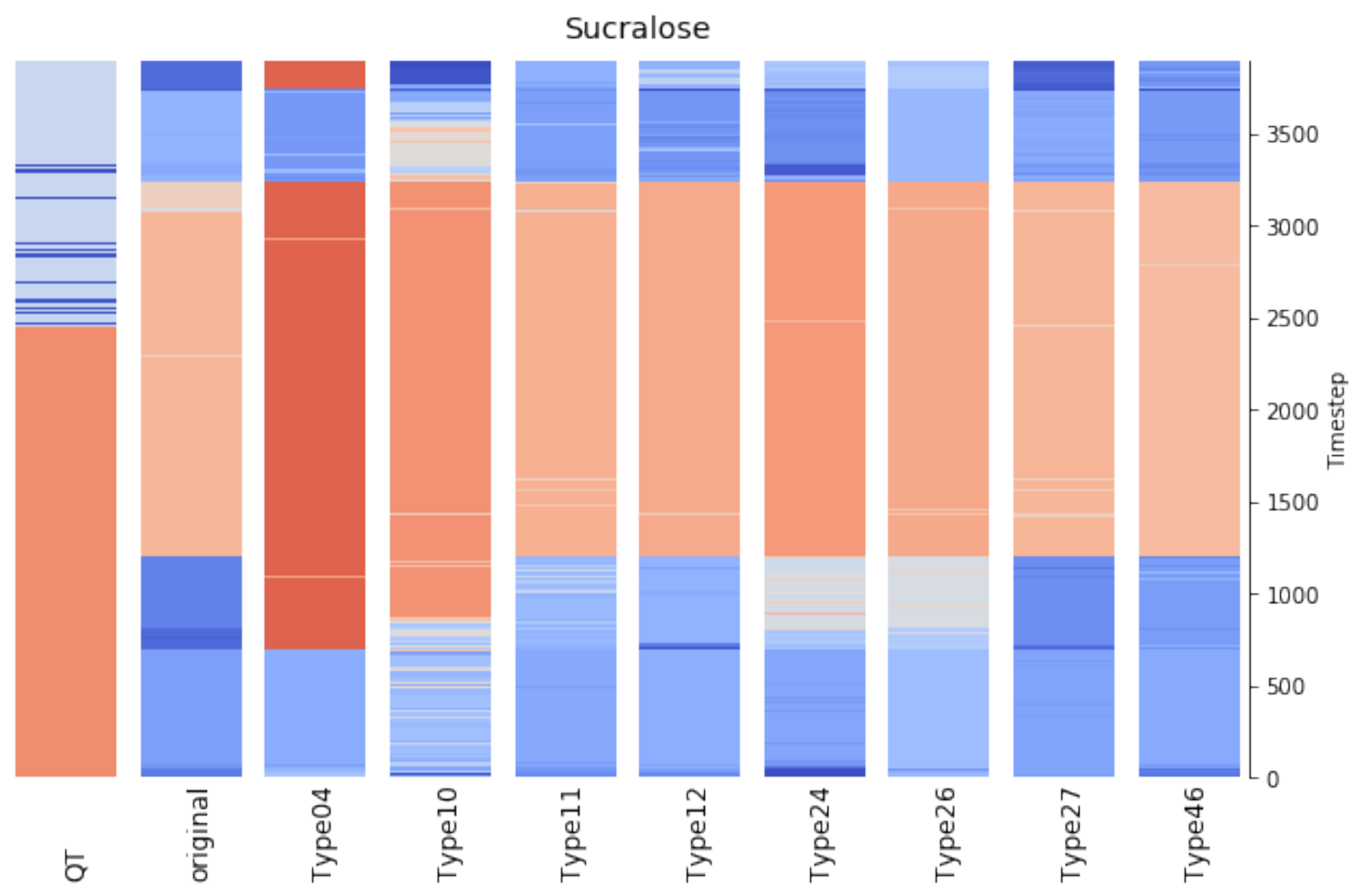}}
\subfloat{\includegraphics[width=0.33\textwidth]{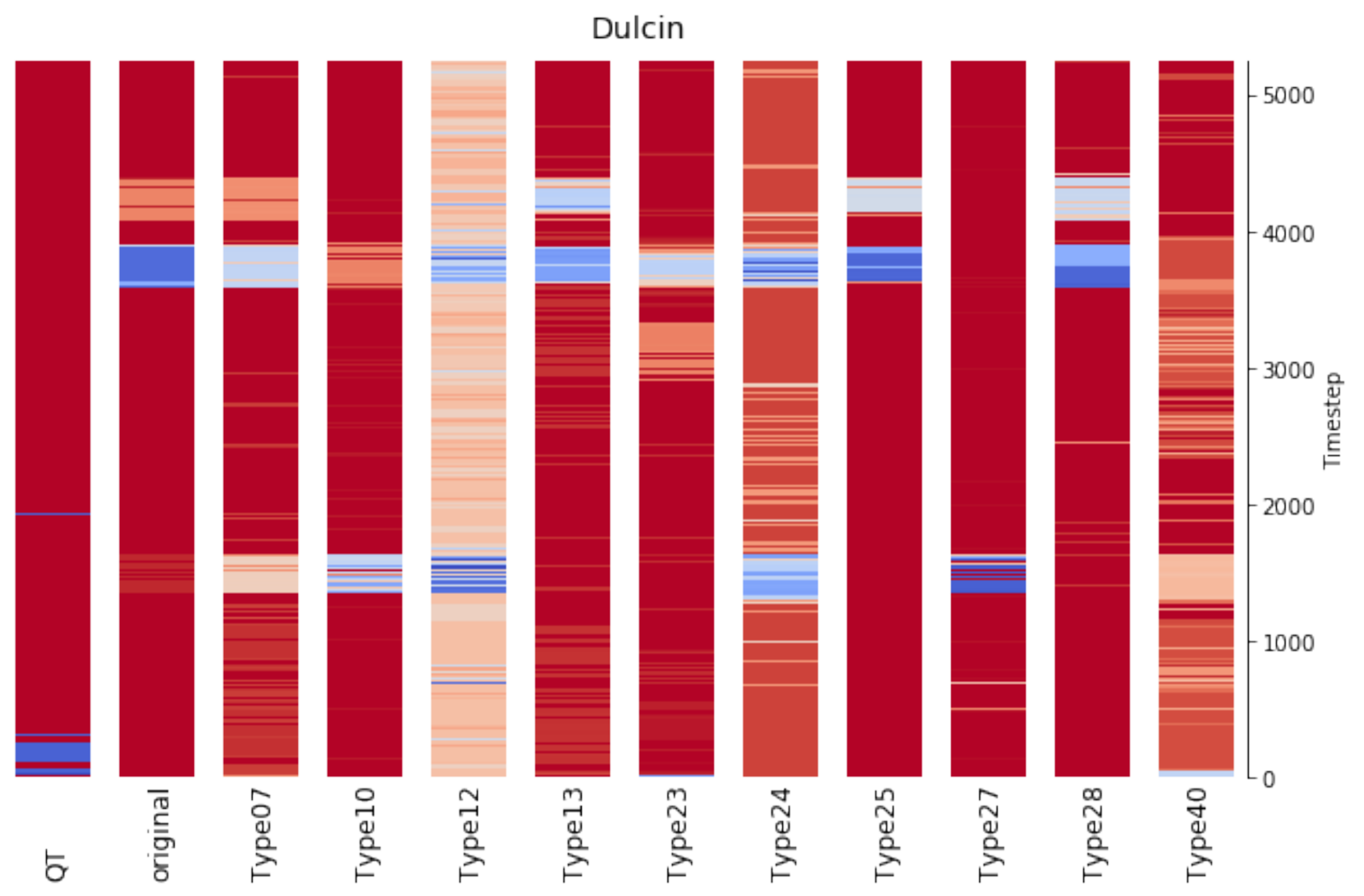}} 
\subfloat{\includegraphics[width=0.33\textwidth]{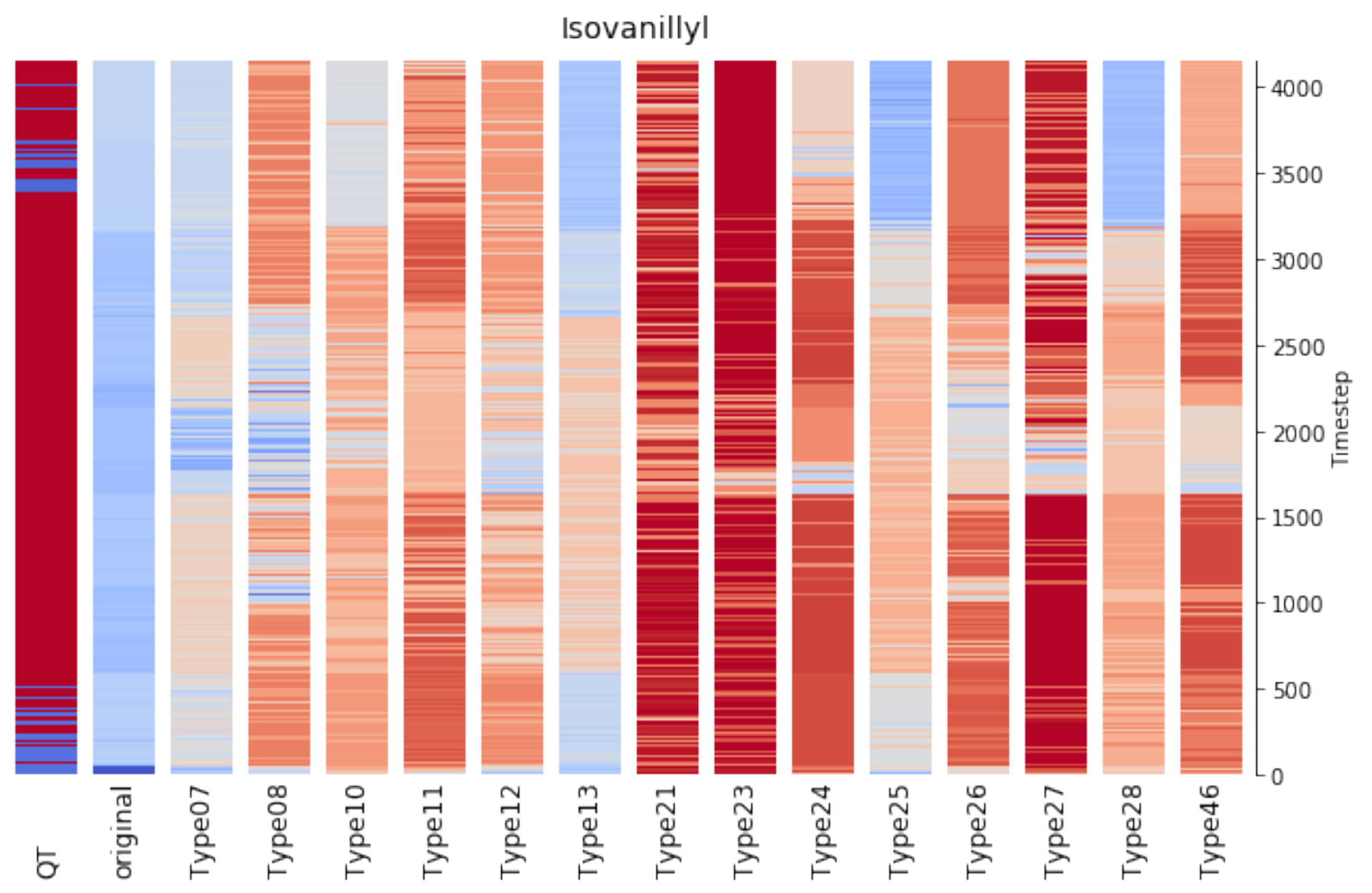}} 
\caption{Clustering results for all atom subtypes in the five sweeteners. The first bar shows the RMSD clustering with VMD1.9.3. The second bar is calculated with CVAE and HDBSCAN when all atoms are considered. The following bars represent the clustering results for each atom subtype of the corresponding sweetener. Each plot depicts the cluster size at time $t$ against simulation time $t$. For CVAE and HDSCAN results, each cluster size $t$ is the average of the sizes given by the $A$ models. Bigger clusters are colored with dark red while small clusters are colored with dark blue.}
\label{fig:clusters_comparison}
\end{figure}

We further analyzed the behavior of each atom subtype and compared it to the RMSD reference. 
%Figure \ref{fig:clusters_comparison} shows the clustering comparison for the five sweeteners. 
With the first bar as reference from RMSD, Figure \ref{fig:clusters_comparison} offers a visual comparison of the differences in clustering. In particular, each bar was created with a color map that shows the cluster size of each time step $t$ of the simulation (with bigger clusters in dark red and smaller in dark blue). The second bar shows the clusters obtained by CVAE and HDBSCAN when using all the sweetener atoms. The remaining bars show the clustering results for each individual subtype. For all the AI-enhanced bars, the averaged cluster size among the clustering results given by the different model trained is showed. 
All trajectories have a stable binding mode according to RMSD clustering, that showed big red clusters along almost all the simulations. Only sucralose showed a less stable behavior, especially in the final part of the trajectory. 
As a confirmation of the agreement between RMSD and AI-enhanced clustering, the first two bars of each sweetener agreed well. Only for isovanillyl this agreement was less marked (as also showed in Figure \ref{fig:tsne}).
The different binding behaviors of the individual subtypes were observable by comparing the subtypes cluster sizes with the RMSD ones. Reddish bars identified stable subtypes, for instance subtype 11 and 27 showed a stable result for sucrose. 
Inversely, bars with a different coloring with respect to the reference, hinted less stable subtypes that could be marked as "malicious atoms", because they might weaken the protein-sweetener interactions. In order to quantitatively identify the malicious atoms, we adopted comparison metrics, as reported below.

\subsection*{Comparison metrics such as cosine similarity pointed out the malicious atoms.}
\begin{figure}[h]
\centering
\includegraphics[width=0.4\textwidth]{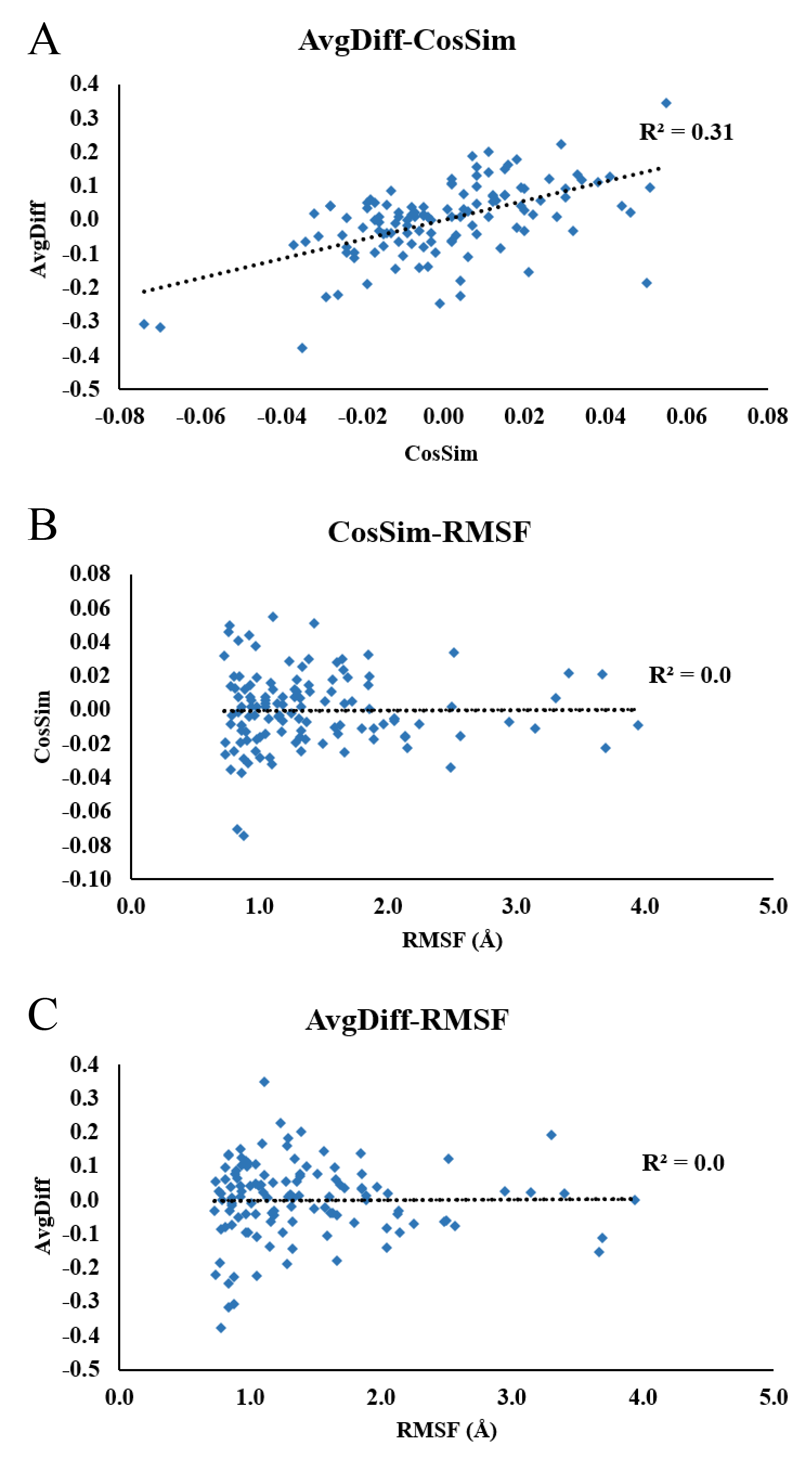}
\caption{Comparison metrics are calculated by comparing the clusters with conventional RMSD clustering and AI-enhanced clusering with Equations \ref{eq:cossim} and \ref{eq:avgdiff}. In (A), we show that the two comparison metrics CosSim and AvgDiff are positively correlated. In (B) and (C), we calculated the RMSF of the atom subtypes for all T1R2-sweetener trajectories. Clearly, CosSim and AvgDiff are not correlated with RMSF.}
\label{fig:corr}
\end{figure}

If the sweeteners remained stable in T1R2, the atom subtypes that deviated from the overall stability should be the "malicious atoms". CASTELO was designed to process MD simulations with stable binding modes to find these malicious atoms that weakened T1R2-sweetener binding affinity. We used Equations \ref{eq:cossim} and \ref{eq:avgdiff} to compare the stability between the whole molecule (calculated with RMSD clustering) and each of the atom subtypes (calculated with AI-enhanced clustering). The resulting CosSim and AvgDiff metrics were plotted in Figure \ref{fig:corr}A from all simulations. It was not surprising that these two metrics were positively correlated (R$^2$ = 0.31).

It is important to mention that the deviation between atom subtypes and the whole molecule should not be comprehended as a time-averaged measurement such as root mean square fluctuation (RMSF). If we consider an atom subtype that only fluctuates drastically when the whole molecule is unstable, the RMSF of this atom subtype might be high but it does not deviate from the overall behavior of the molecule. Therefore, each of the timeframes must be compared separately between the whole molecule and its atom subtypes. With this in mind, Equations \ref{eq:cossim} and \ref{eq:avgdiff} are used as the comparison metrics. Another advantage of AI-enhanced clusters is that they contain rotational and vibrational information, compared to the conventional RMSF calculations that mainly consider translational fluctuations. The additional rotational information from AI-enhanced clusters might be particularly useful to monitor atom groups such as -CH$_3$ and -CH$_2$- that would have been easily ignored otherwise. To clearly show that our metrics differ from conventional RMSF calculations, we compared the CosSim and AvgDiff to RMSF values from all simulations in Figure \ref{fig:corr}B and \ref{fig:corr}C. Both plots yielded R$^2$ = 0.0, indicating that CosSim and AvgDiff contained independent information from RMSF. 

\begin{figure}[h]
\centering
\includegraphics[width=0.9\textwidth]{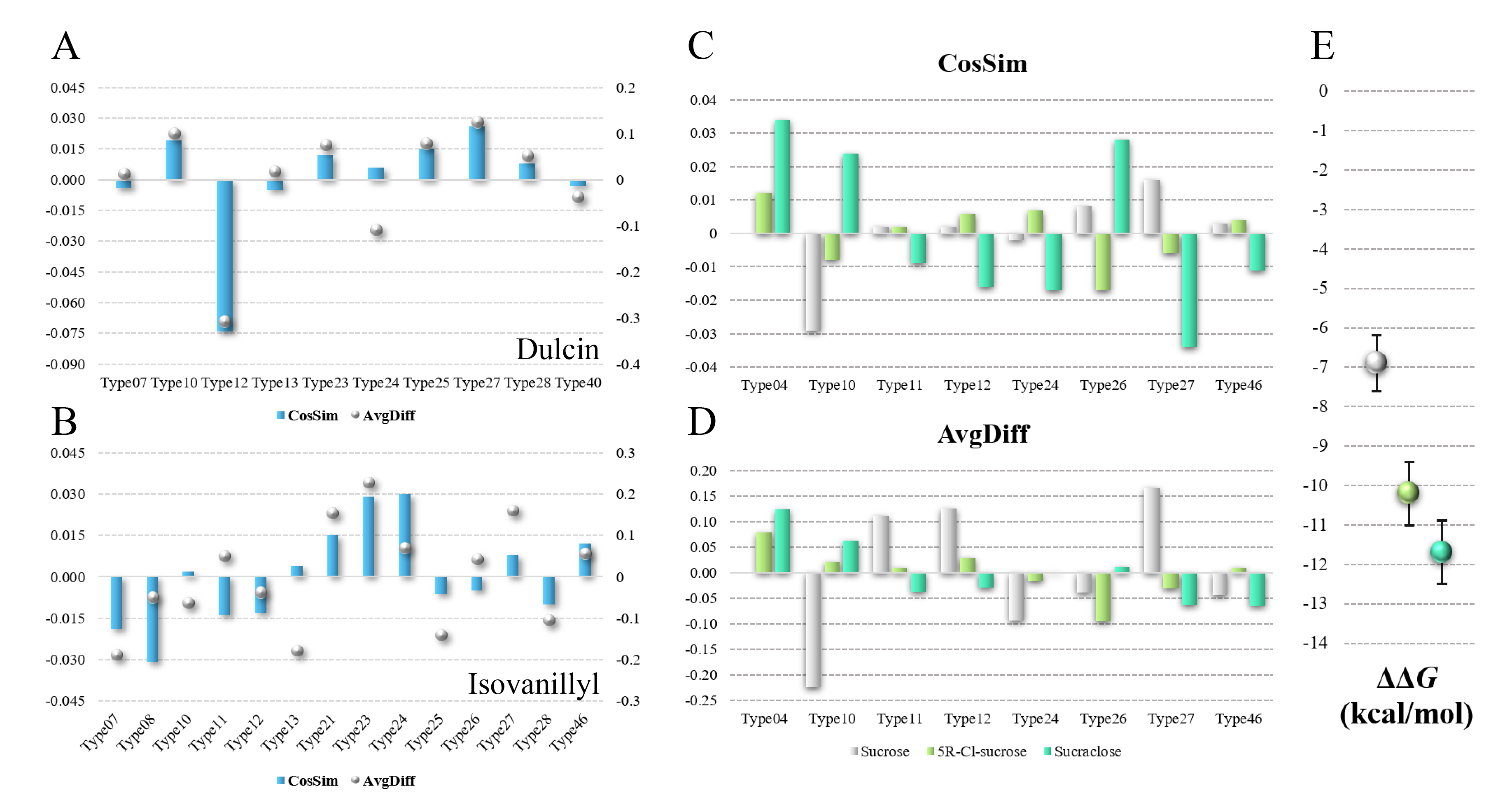}
\caption{We rank the atom subtypes from the five sweeteners with the calculated comparison metrics CosSim and AvgDiff. Negative values mean that the corresponding atoms are "malicious". Positive values mean that the corresponding atoms are "beneficial". The results for dulcin and isovanillyl are plotted separately in (A) and (B). We group sucrose, 4R-Cl sucrose and sucralose due to their structural similarity. The CosSim and AvgDiff values of the 8 atom subtypes shared among these three sweeteners are plotted in (C) and (D), respectively. The relative binding free energy (E, with ethane as the reference point) is plotted in juxtaposition to the comparison metrics.}
\label{fig:leadopt}
\end{figure}

To find the malicious atoms, we focused on the trajectories that comprised largest clusters, such as the ones shown in Figure \ref{fig:clusters_comparison}. CosSim and AvgDiff metrics were plotted in Figure \ref{fig:leadopt} for these trajectories. A ranking system was used to pick out the atom subtype most prone to undermine the overall T1R2-sweetener binding stability. For example, in Figure \ref{fig:leadopt}A, subtype 12 of dulcin was clearly the malicious atom subtype, because it ranked the lowest in both CosSim and AvgDiff metrics. In Figure \ref{fig:leadopt}B, subtype 7 and 8 were mostly likely the malicious atom subtypes, because of their low values in both CosSim and AvgDiff. The selected atom subtypes were chosen as the candidate for lead optimization (see results in the next section).

In Figure \ref{fig:leadopt}C and \ref{fig:leadopt}D, we compared the comparison metrics of sucrose, 4R-Cl-sucrose and sucralose, which could be considered as a lead optimization chain due to their structural similarity. We noticed that the added chlorine atoms (subtype 4) in 4R-Cl-sucrose and sucralose were increasingly beneficial to the overall structural stability (both CosSim and AvgDiff metrics were increasingly positive). The suggested malicious atom subtypes were carbohydrate carbons that connect with hydroxyl groups (-CHOH-, subtype 10). Upon modifying several of the hydroxyl groups to chlorine atoms, both CosSim and AvgDiff metrics of subtype 10 were increasingly more positive, indicating that the replacement of chlorine atoms affected the stability of its connecting carbons. As a comparison, we plotted the relative binding free energy of sucrose, 4R-Cl-sucrose and sucralose in Figure \ref{fig:leadopt}E, which clearly correlated with CosSim and AvgDiff metrics for subtype 4 and subtype 10. Finally, we provided a possibility to further improve sucralose by suggesting that the malicious atoms in sucralose were probably from subtype 27, which were the aliphatic hydrogens in the -CH$_2$OH group. %It is interesting that the comparison metrics of hydroxyl groups (subtype 46) remained neutral for all 3 sweeteners, suggesting either further optimization opportunities or a limitation of our method. 

\subsection*{FEP calculations verified that malicious atom types could be modified to strengthen T1R2-sweetener binding affinity, providing an opportunity for lead optimization. }
\begin{figure}[h]
\centering
\includegraphics[width=0.9\textwidth]{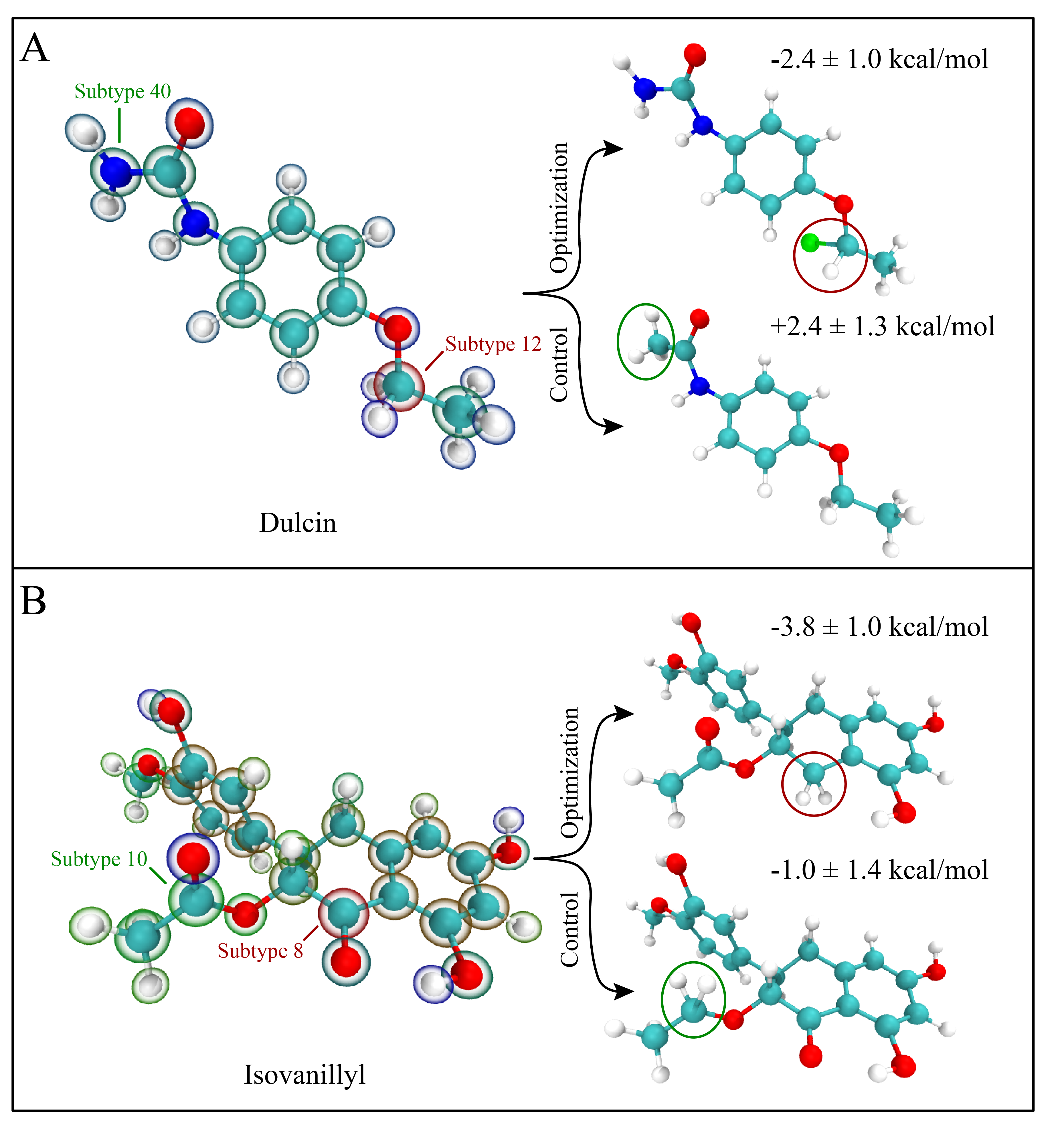}
\caption{Attempted lead optimization for dulcin (A) and isovanillyl (B). The molecules are illustrated with two representations: solid stick and ball model for atom identification (cyan for carbon, white for hydrogen, blue for nitrogen, red for oxygen and green for chlorine); bubbles for CosSim values (blue for positive/beneficial, green for neutral and red for negative/malicious). Two examples are illustrated, one for optimization and one as control, for both dulcin and isovanillyl. The resulting binding free energy changes are listed next to the examples, with negative values meaning strengthened binding affinity and positive values meaning weakened binding affinity. A complete list of the lead optimization can be found in Table \ref{tab:dulcin} and Table \ref{tab:isov}.}
\label{fig:mod}
\end{figure}

Based on the ranked comparison metric values in Figure \ref{fig:leadopt}A and \ref{fig:leadopt}B, we visualized the malicious atoms of dulcin and isovanillyl in Figure \ref{fig:mod}A and \ref{fig:mod}B, as a way to guide chemists in lead optimization. For example, subtype 12 of dulcin could be easily changed to a carbonyl (-CO-) or a chloronated carbon (-CHCl-). Subtype 8 of isovanillyl could be easily changed to an aliphatic carbon (-CH$_2$-). In addition to these malicious atoms, we also selected some neutral atoms as control groups (like subtype 40 of dulcin and subtype 10 of isovanillyl). To verify if the identified atoms were indeed "malicious", FEP calculations were adopted to compare the binding free energy of the modified lead molecules to the lead molecules (exemplified in Figure \ref{fig:mod}; a full list can be found in Table \ref{tab:dulcin} and Table \ref{tab:isov}). For example, a dual topology was made between dulcin and modified dulcin where group X (also subtype 12, see Table \ref{tab:dulcin}) was changed from CH$_2$ to CHCl. Two sets of simulations were performed: one in T1R2 and one solvated with 0.1 M NaCl solution. The calculated binding free energy (${\Delta}{\Delta}F$) was thus a direct comparison between dulcin and the modified dulcin. 

We tried 4 combinations for dulcin optimization varying neutral moiety R and malicious moiety X, shown in Table \ref{tab:dulcin}. Modifications on moiety R always weakened T1R2-dulcin binding affinity. However, modifications on moiety X resulted in one favorable modifications. -2.4 $\pm$ 1.0 kcal/mol was obtained when moiety X was changed from -CH$_2$- to -CHCl-. Interestingly, when moiety X was changed to -CO-, the binding affinity was weakened, probably because its surroundings was slightly hydrophobic (I327, seen in Figure \ref{fig:structures}E).

Similarly, we attempted 2 combinations for isovanillyl optimization, varying neutral moiety R and malicious moiety X, shown in Table \ref{tab:isov}. Modifications on moiety R neither strengthened nor weakened T1R2-isovanillyl binding affinity (${\Delta}{\Delta}F$ was not significant compared to its standard deviation). In contrast, modifications on moiety X significantly strengthened T1R2-isovanilly binding affinity by -3.8 $\pm$ 1.0 kcal/mol. It was counter-intuitive that changing moiety X to a hydrophobic group would significantly benefit the binding affinity, given that its surroundings included D142 and Y164 (Figure \ref{fig:structures}F). On the other hand, the surroundings of moiety R included V388, F39 and L71, which would likely favor a hydrophobic modification. Further analysis pointed out that by modifying moiety X, its nearby phenol-hydroxyl group was freed from forming intramolecular hydrogen bond and established a hydrogen bond with D142. This transition lowered the electrostatic penalty of charged D142 residue around the hydrophobic isovanillyl (evidenced in Table \ref{tab:fedecomp}). Our findings blindly agreed with a previous isovanillyl optimization which suggested the optimal moiety X to be -CH$_2$-, \cite{bassoli2002isovanillyl} reinforcing that CASTELO could potentially assist in lead optimization.

\section*{Conclusions}
In this work, we introduced CASTELO, a ML-MD pipeline that processed MD simulation data with drug targets and their known leads and suggested modifiable submolecular moieties for lead optimization. We generated dynamism tensors by including temporal information in the conventional contact matrices. CVAE method was adopted to compress the dynamism tensors into latent space before the data clustering with HDBSCAN. The lead molecule was grouped into atom subtypes to pin down submolecular contributions to the target-lead binding affinity. The resulting cluster information was compared to a traditional RMSD-based clustering method for the whole molecule with the proposed comparison metrics CosSim and AvgDiff. Finally, we ranked the submolecular moieties to find clues for lead optimization. With T1R2-sweetener as a model system, we proved that our pipeline nicely explained the improvement of the sweetness from sucrose to 4R-Cl-sucrose and sucralose. Most notably, we suggested two brand new molecules based on the CASTELO pipeline using T1R2-dulcin and T1R2-isovanillyl simulations. With free energy calculations, we verified that the newly improved dulcin was $\sim$57 times sweeter than dulcin, $\sim$14,000 times sweeter than sucrose. The newly improved isovanillyl was computed to be $\sim$600 times sweeter than isovanillyl, 240,000 times sweeter than sucrose. Our physics-based model should be transferable to other systems, given that newest development of the force field could be trusted. We plan to use more target-lead systems to test CASTELO's scalability and interpretability. For example, a similar approach could be adapted for major histocompatibility complex (MHC) and epitope complexes (paper submitted). The identification of any moiety may indicate destabilizing motions of its surrounding atoms rather than moiety itself. Generative models could be developed on top of the malicious atom identifications in this study. Although CASTELO's dependency on experimental data is low, we foresee that our tool could be further enhanced with more use cases, especially with more experimental data to reference to.

\section*{Supplementary material}
Table \ref{tab:fedecomp} provides the binding free energy decomposition on the T1R2-sweetener systems. Table \ref{tab:dulcin} is provided for the \textit{in silico} lead optimization of dulcin. Table \ref{tab:isov} is provided for the \textit{in silico} lead optimization of isovanillyl. Figure \ref{fig:thermocycle} is attached to show the thermodynamic cycle used for FEP calculations. Figure \ref{fig:clussize} plots the stable clusters identified by RMSD clustering. Figure \ref{fig:structures} illustrates the stable binding structures for the five sweeteners in the T1R2 flytrap domain.

%\section*{Competing interests}
%The proposed pipeline has been submitted to USPTO in October, 2020 under the patent name of Altering Protein-Ligand Structure According to Protein-Ligand Binding Affinity, by authors: Giacomo  Domeniconi, Leili  Zhang, Guojing  Cong, Chih-Chieh  Yang and Ruhong Zhou.

\section*{Author's contributions}
L.Z. and G.D. conceived of the presented work. L.Z. and S.K. performed the MD/FEP simulations. L.Z. and G.D. extracted and processed contact data from the simulations. G.D. and C.Y. trained the CVAE models. L.Z. and G.D. conceived of the comparison metrics for the final ranking of submolecular moieties. R.Z. and G.C. supervised the findings of this work. All authors discussed the results and contributed to the final manuscript.

\section*{Acknowledgements}
We thank Josef Klucik and Paul Winget for valuable discussions on the subject of the sweeteners. R.Z. and G.C. gratefully acknowledge the financial support from the IBM Bluegene Science Program (W125859, W1464125 and W1464164), Computing Cloud Clusters and Witherspoon supercomputer in IBM.
%%  After submission of the .TEX file,                     %%
%%  you will be prompted to submit your .BBL file.         %%

% if your bibliography is in bibtex format, use those commands:
%\bibliographystyle{bmc-mathphys} % Style BST file
\bibliographystyle{plain} % Style BST file
\bibliography{bmc_article}      % Bibliography file (usually '*.bib' )

\pagebreak
\begin{center}
\textbf{\huge{Supplementary Material}} \\
\  \\
for \\
\  \\
\textbf{\large{CASTELO: Clustered Atom Subtypes aidEd Lead Optimization – a combined machine learning and molecular modeling method}}
\end{center}

        \setcounter{table}{0}
        \renewcommand{\thetable}{S\arabic{table}}
        \setcounter{figure}{0}
        \renewcommand{\thefigure}{S\arabic{figure}}

\begin{table}[h]
\centering
\begin{tabular}{l | c | c | c}
\hline
Sweetener & ${\Delta\Delta}F$ & ${\Delta\Delta}F_{elec}$ & ${\Delta\Delta}F_{vdw}$ \\
\hline
Sucrose & -6.9 ${\pm}$ 0.7 & -8.0 ${\pm}$ 0.5 & 0.2 ${\pm}$ 0.5 \\
4R-Cl-sucrose & -10.2 ${\pm}$ 0.8 & -9.3 ${\pm}$ 0.5 & -2.3 ${\pm}$ 0.6 \\
Sucralose & -11.7 ${\pm}$ 0.8 & -11.5 ${\pm}$ 0.6 & -1.9 ${\pm}$ 0.6 \\
Dulcin & -10.6 ${\pm}$ 0.4 & -7.1 ${\pm}$ 0.3 & -3.2 ${\pm}$ 0.6 \\
Isovanillyl & -11.1 ${\pm}$ 0.8 & 0.8 ${\pm}$ 0.5 & -12.8 ${\pm}$ 0.6 \\
\hline
\end{tabular}
\caption{Binding free energy decomposition. ${\Delta\Delta}F_{elec}$ or ${\Delta\Delta}F_{vdw}$ is calculated with Zwanzig equation when only electrostatic energy or VDW energy is considered. Note that due to the construction of Zwanzig equation, ${\Delta\Delta}F$ is not a simple addition of ${\Delta\Delta}F_{elec}$ or ${\Delta\Delta}F_{vdw}$.}
\label{tab:fedecomp}
\end{table}

\begin{table}[h]
\centering
\raisebox{-0.3\totalheight}{\includegraphics[width=0.3\textwidth]{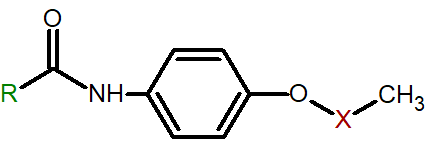}}
\begin{tabular}{l | l | c }
\hline
R & X & ${\Delta\Delta}F$ \\
\hline
NH$_2$ & CH$_2$                 & 0 \\
NH$_2$ & CO                        & +2.5 ${\pm}$ 1.1 \\
NH$_2$ & CHCl                     & -2.4 ${\pm}$ 1.0 \\
CH$_3$ & CH$_2$                 & +2.4 ${\pm}$ 1.3 \\
NHCH$_2$CH$_3$ & CH$_2$ & +4.9 ${\pm}$ 1.3 \\
\hline
\end{tabular}
\caption{Lead optimization for dulcin. Relative free energy ${\Delta\Delta}F$ is calculated using the original dulcin molecule as the reference (where R is NH$_2$ and X is CH$_2$). Negative values of ${\Delta\Delta}F$ indicate that the modifications strengthen the binding affinity. The unit of ${\Delta\Delta}F$ is kcal/mol.}
\label{tab:dulcin}
\end{table}

\begin{table}[h]
\centering
\raisebox{-0.5\totalheight}{\includegraphics[width=0.3\textwidth]{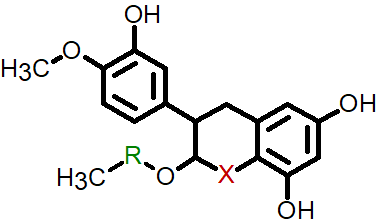}}
\begin{tabular}{l | l | c }
\hline
R & X & ${\Delta\Delta}F$ \\
\hline
CO        & CO                 & 0 \\
CO        & CH$_2$          & -3.8 ${\pm}$ 1.0 \\
CH$_2$ & CO                 & -1.0 ${\pm}$ 1.4 \\
\hline
\end{tabular}
\caption{Lead optimization for isovanillyl. Relative free energy ${\Delta\Delta}F$ is calculated using the original isovanillyl molecule as the reference (where R and X are both CO). Negative values of ${\Delta\Delta}F$ indicate that the modifications strengthen the binding affinity. The unit of ${\Delta\Delta}F$ is kcal/mol.}
\label{tab:isov}
\end{table}

\begin{figure}[h]
\centering
\includegraphics[width=0.5\textwidth]{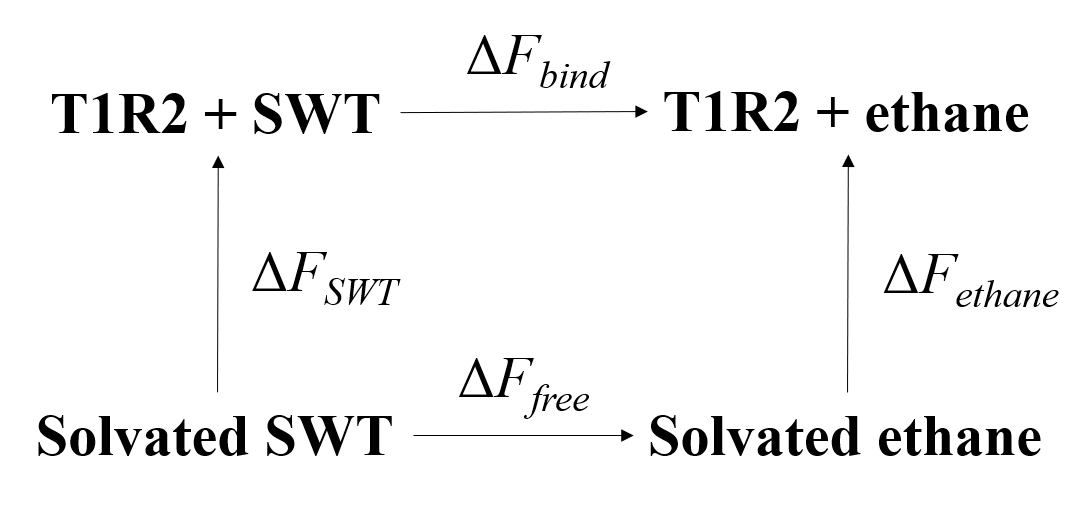}
\caption{The thermodynamic cycle for FEP calculations. See method for details.}
\label{fig:thermocycle}
\end{figure}

\pagebreak
\begin{figure}[h]
\centering
\includegraphics[width=0.6\textwidth]{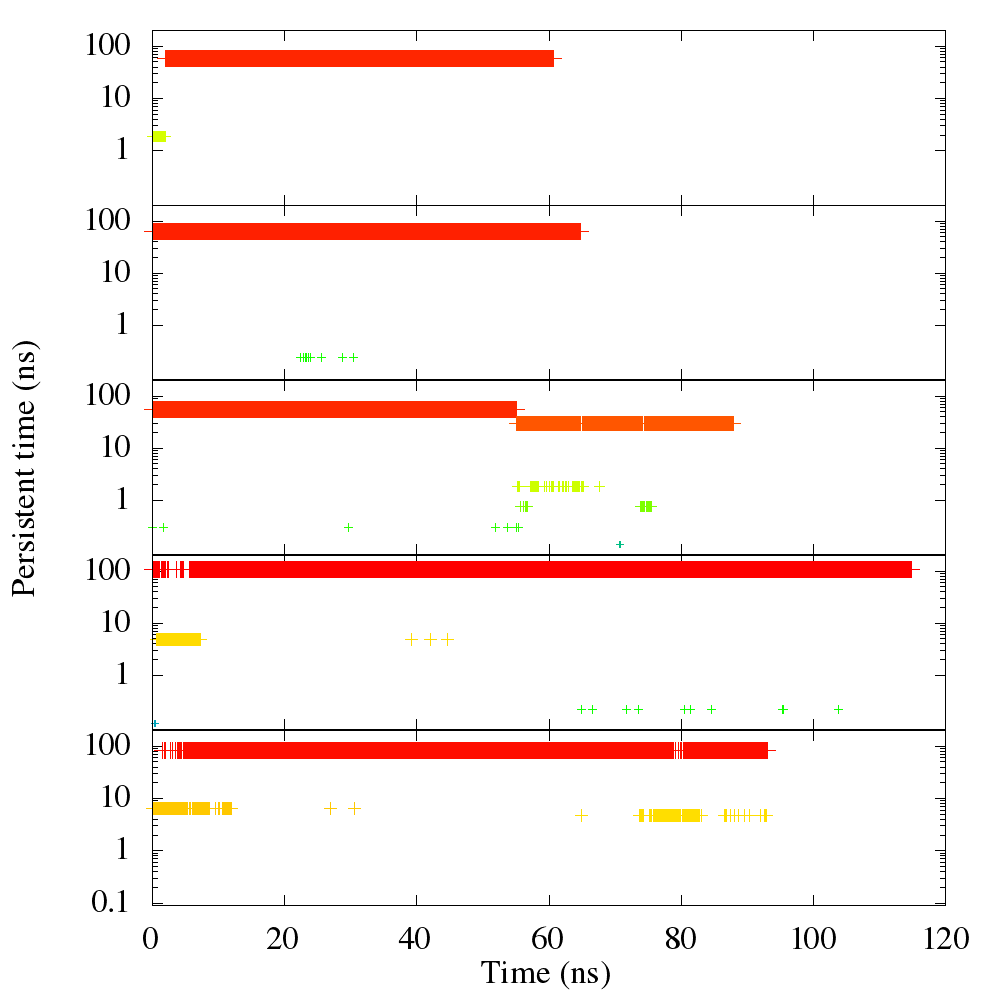}
\caption{From the MD simulations, we identify the clusters of sweetener binding states with RMSD calculations on the sweetener alone, while the structure of T1R2 is aligned. The cutoff for the RMSD clustering method is 2 {\AA}. The clusters are plotted using their "persistent time" (defined as the total time each cluster persists in the corresponding simulation). From top to bottom, we plot the persistent time of sucrose, 4R-Cl-sucrose, sucralose, dulcin and isovanillyl, respectively.}
\label{fig:clussize}
\end{figure}

\pagebreak
\begin{figure}[h]
\centering
\includegraphics[width=0.9\textwidth]{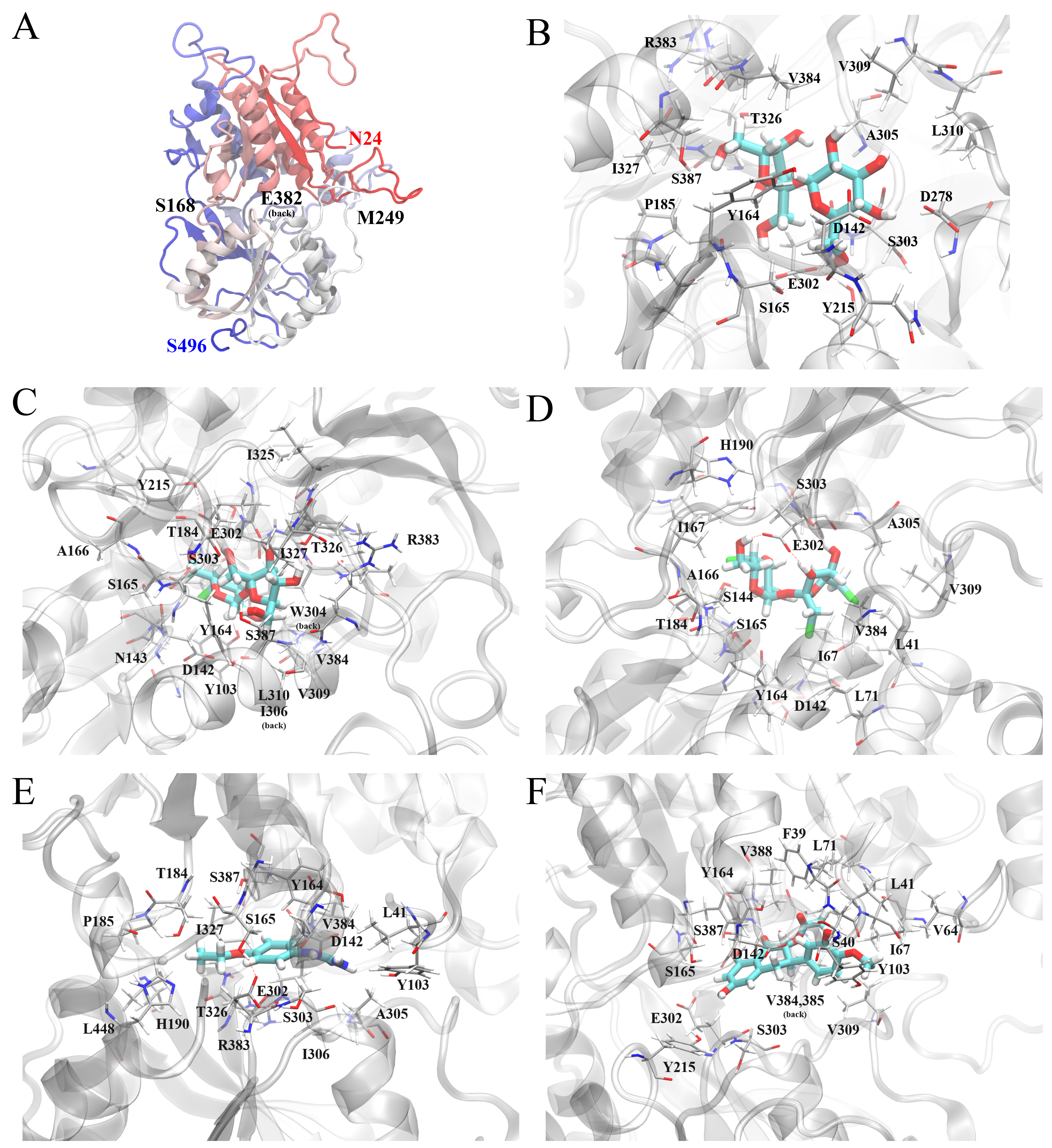}
\caption{(A) We illustrate T1R2 protein structure with N-terminus colored by red and C-terminus colored by blue. Stable binding structures of the five sweeteners in T1R2 are identified from the MD simulations and drawn in (B) with sucrose, (C) with 4R-Cl-sucrose, (D) with sucralose, (E) with dulcin and (F) with isovanillyl, respectively. Protein structures are drawn with transparent "newcartoon" representation in VMD. Sweeteners and their interacting residues are drawn with sticks models. The colors for the sticks models are selected as follows: cyan for carbon in the sweeteners, grey for carbon in T1R2, white for hydrogen, red for oxygen, blue for nitrogen, and green for chlorine.}
\label{fig:structures}
\end{figure}

%\end{backmatter}

\end{document}